%% file: main.tex
\documentclass[11pt, a4paper, logo, copyright]{hail_lab}

\usepackage[utf8]{inputenc} 
\usepackage[T1]{fontenc}    
\usepackage{url}            
\usepackage{booktabs}       
\usepackage{amsfonts}       
\usepackage{nicefrac}       
\usepackage{microtype}      

\input{math_commands.tex}
\usepackage[utf8]{inputenc}
\usepackage[T1]{fontenc}
\usepackage{microtype}

\usepackage{amsmath,amssymb,amsfonts,bm}
\usepackage{amsthm}
\usepackage{mathtools}
\usepackage{bbm}

\usepackage{booktabs}
\usepackage{array}
\usepackage{tabularx}
\usepackage{ragged2e}
\usepackage{makecell}
\usepackage{multirow}
\usepackage{colortbl}
\usepackage{longtable}
\usepackage{adjustbox}

\usepackage{graphicx}
\usepackage{subcaption}
\usepackage{wrapfig}
\usepackage{float}

\usepackage{xcolor}
\colorlet{darkgreen}{green!65!black}
\colorlet{darkblue}{blue!75!black}
\colorlet{darkred}{red!80!black}
\definecolor{statistical}{HTML}{8c564b}
\definecolor{structural}{HTML}{0070C0}
\definecolor{semantic}{HTML}{008080}
\definecolor{yellow}{HTML}{f7c600}
\definecolor{lightblue}{HTML}{0071bc}
\definecolor{lightgreen}{HTML}{39b54a}
\definecolor{mypurple}{HTML}{412F8A}
\definecolor{myorange}{HTML}{fc8e62}
\definecolor{textgreen}{RGB}{57, 172, 57}
\definecolor{textred}{RGB}{200, 10, 10}
\definecolor{deemph}{gray}{0.55}
\definecolor{baselinecolor}{gray}{.95}
\definecolor{graycolor}{gray}{.95}

\definecolor{accel_color}{rgb}{0.13, 0.25, 0.42}
\definecolor{gyro_color}{rgb}{0.28, 0.46, 0.64}
\definecolor{mag_color}{rgb}{0.50, 0.65, 0.80}
\definecolor{wrist_color}{rgb}{0.48, 0.12, 0.00}
\definecolor{hand_color}{rgb}{0.57, 0.18, 0.04}
\definecolor{arm_color}{rgb}{0.66, 0.25, 0.10}
\definecolor{back_color}{rgb}{0.74, 0.33, 0.18}
\definecolor{chest_color}{rgb}{0.80, 0.40, 0.25}
\definecolor{hip_color}{rgb}{0.85, 0.50, 0.28}
\definecolor{knee_color}{rgb}{0.88, 0.55, 0.36}
\definecolor{leg_color}{rgb}{0.91, 0.60, 0.44}
\definecolor{shank_color}{rgb}{0.94, 0.66, 0.52}
\definecolor{ankle_color}{rgb}{0.95, 0.70, 0.60}

\newcommand{\badge}[2]{\colorbox{#1}{\textcolor{white}{\texttt{#2}}}}

\newcommand{\bACC}{\badge{accel_color}{ACC}}
\newcommand{\bGYR}{\badge{gyro_color}{GYR}}
\newcommand{\bMAG}{\badge{mag_color}{MAG}}

\newcommand{\bWR}{\badge{wrist_color}{WR}}
\newcommand{\bHN}{\badge{hand_color}{HN}}
\newcommand{\bAR}{\badge{arm_color}{AR}}
\newcommand{\bBK}{\badge{back_color}{BK}}
\newcommand{\bCH}{\badge{chest_color}{CH}}
\newcommand{\bHP}{\badge{hip_color}{HP}}
\newcommand{\bKN}{\badge{knee_color}{KN}}
\newcommand{\bLG}{\badge{leg_color}{LG}}
\newcommand{\bSH}{\badge{shank_color}{SH}}
\newcommand{\bAN}{\badge{ankle_color}{AN}}
\newcommand{\wrapitem}[1]{%
  \par
  \begingroup
  \noindent
  \leftskip=1.3em
  \parindent=-1.3em
  \llap{\textbullet\hspace{0.5em}}#1\par
  \endgroup
}

\usepackage[pagebackref=false, breaklinks=true, colorlinks, citecolor=lightblue, urlcolor=lightblue, bookmarks=false]{hyperref}
\usepackage{url}
\usepackage[capitalize,noabbrev]{cleveref}
\usepackage{cite}

\usepackage{enumitem}

\usepackage{xspace}
\usepackage{pbox}
\usepackage{nicefrac}
\usepackage{afterpage}
\usepackage{textcomp}
\usepackage{siunitx}
\usepackage{soul}
\usepackage{etoolbox}
\usepackage[most]{tcolorbox}
\tcbset{
    frame code={},
    center title,
    left=0pt,
    right=0pt,
    top=0pt,
    bottom=0pt,
    colframe=white,
    width=\dimexpr\textwidth\relax,
    enlarge left by=0mm,
    boxsep=5pt,
    arc=0pt,outer arc=0pt,
}
\usepackage{pifont}
\usepackage{bbding}
\definecolor{datascaling}{HTML}{D76700}
\definecolor{modelscaling}{HTML}{1A72C9}
\usepackage{listings}

\newcolumntype{L}[1]{>{\RaggedRight\arraybackslash}p{#1}}
\newcolumntype{Y}{>{\raggedright\arraybackslash}X}
\newcolumntype{x}[1]{>{\centering\arraybackslash}p{#1pt}}
\newcolumntype{y}[1]{>{\raggedright\arraybackslash}p{#1pt}}
\newcolumntype{z}[1]{>{\raggedleft\arraybackslash}p{#1pt}}

\newlength\savewidth
\newcommand{\grayrow}{\rowcolor[gray]{.95}}
\newcommand{\graycell}[1]{\cellcolor{baselinecolor}{#1}}

\newcommand{\ryes}{\textcolor{green!60!black}{\ding{51}}}
\newcommand{\rno}{\textcolor{red!75!black}{\ding{55}}}

\newcommand{\tabref}[1]{Table \ref{#1}}

\newcommand{\ours}{\texttt{Inertia-1}\xspace}
\newcommand{\name}{\ours}  
\newcommand{\patchtst}{{PatchTST}\xspace}
\newcommand{\selfpab}{{Self-PAB}\xspace}
\newcommand{\simclr}{{SimCLR}\xspace}
\newcommand{\dino}{{DINO}\xspace}
\newcommand{\argru}{AR$_{\mathrm{G}}$\xspace}
\newcommand{\artransformer}{AR$_{\mathrm{T}}$\xspace}
\newcommand{\lsm}{{LSM}\xspace}
\newcommand{\relcon}{{RelCon}\xspace}
\newcommand{\sslwearables}{{SSL-Wearables}\xspace}

\newcommand{\har}{HAR\xspace}
\newcommand{\fog}{FoG\xspace}
\newcommand{\dd}{Disease Prediction\xspace}

\newcommand{\patchtstCite}{{PatchTST \cite{nie2023patchtst}}\xspace}
\newcommand{\selfpabCite}{{Self-PAB \cite{logacjov2024selfpab}}\xspace}
\newcommand{\simclrCite}{{SimCLR \cite{chen2020simple}}\xspace}
\newcommand{\dinoCite}{{DINO \cite{caron2021emerging}}\xspace}
\newcommand{\argruCite}{{AR$_{\mathrm{G}}$ \cite{chung2014empiricalevaluationgatedrecurrent}}\xspace}
\newcommand{\artransformerCite}{{AR$_{\mathrm{T}}$ \cite{vaswani2023attentionneed}}\xspace}
\newcommand{\lsmCite}{{LSM \cite{narayanswamy2025scaling}}\xspace}
\newcommand{\relconCite}{{RelCon \cite{xu2025relcon}}\xspace}
\newcommand{\sslwearablesCite}{{SSL-Wearables \cite{yuan2024self}}\xspace}

\newcommand{\nhanes}{\texttt{NHANES}\xspace}
\newcommand{\ukb}{\texttt{UK Biobank}\xspace}
\newcommand{\capture}{\texttt{CAPTURE-24}\xspace}
\newcommand{\harplus}{\texttt{HAR70+}\xspace}
\newcommand{\harth}{\texttt{HARTH}\xspace}
\newcommand{\hhar}{\texttt{HHAR}\xspace}
\newcommand{\mhealth}{\texttt{MHEALTH}\xspace}
\newcommand{\opportunity}{\texttt{OPPORTUNITY}\xspace}
\newcommand{\pamap}{\texttt{PAMAP2}\xspace}
\newcommand{\recofit}{\texttt{RecoFit}\xspace}
\newcommand{\wisdm}{\texttt{WISDM}\xspace}
\newcommand{\wear}{\texttt{WEAR}\xspace}
\newcommand{\daphnet}{\texttt{DaphnetFoG}\xspace}
\newcommand{\odayfog}{\texttt{OdayFoG}\xspace}
\newcommand{\fogturning}{\texttt{FoGTurning}\xspace}

\newcommand{\nhanesCite}{\texttt{NHANES} \cite{nhanes2011}\xspace}
\newcommand{\ukbCite}{\texttt{UK Biobank} \cite{doherty2017large}\xspace}
\newcommand{\captureCite}{\texttt{CAPTURE-24} \cite{Chang2021-vh}\xspace}
\newcommand{\harplusCite}{\texttt{HAR70+} \cite{har70+_780}\xspace}
\newcommand{\harthCite}{\texttt{HARTH} \cite{harth_779}\xspace}
\newcommand{\hharCite}{\texttt{HHAR} \cite{heterogeneity_activity_recognition_344}\xspace}
\newcommand{\mhealthCite}{\texttt{MHEALTH} \cite{mhealth_319}\xspace}
\newcommand{\opportunityCite}{\texttt{OPPORTUNITY} \cite{opportunity_activity_recognition_226}\xspace}
\newcommand{\pamapCite}{\texttt{PAMAP2} \cite{pamap2_physical_activity_monitoring_231}\xspace}
\newcommand{\recofitCite}{\texttt{RecoFit} \cite{Morris_Saponas_Guillory_Kelner_2014}\xspace}
\newcommand{\wisdmCite}{\texttt{WISDM} \cite{wisdm_smartphone_and_smartwatch_activity_and_biometrics_dataset__507}\xspace}
\newcommand{\wearCite}{\texttt{WEAR} \cite{bock2024wearoutdoorsportsdataset}\xspace}
\newcommand{\daphnetCite}{\texttt{Daphnet FoG} \cite{daphnet_freezing_of_gait_245}\xspace}
\newcommand{\odayfogCite}{\texttt{OdayFoG} \cite{o2022assessing}\xspace}
\newcommand{\fogturningCite}{\texttt{FoGTurning} \cite{10.3389/fnins.2022.832463}\xspace}

\theoremstyle{plain}

\theoremstyle{definition}

\theoremstyle{remark}

\usepackage{titletoc}
\usepackage{letltxmacro}

\definecolor{textgreen}{RGB}{57, 172, 57}
\definecolor{textred}{RGB}{200, 10, 10}
\definecolor{boxyellow}{HTML}{FAF5E6}
\definecolor{frameyellow}{HTML}{B7950B}
\definecolor{boxpurple}{HTML}{F4EFF6}
\definecolor{framepurple}{HTML}{6C3483}
\definecolor{boxblue}{HTML}{EEF4F8}
\definecolor{frameblue}{HTML}{2874A6}
\definecolor{boxgray}{HTML}{F0F2F3}
\definecolor{framegray}{HTML}{5D6D7E}
\definecolor{boxgreen}{HTML}{EAFaf1}
\definecolor{framegreen}{HTML}{196F3D}
\usepackage{pifont} 

\usepackage[most]{tcolorbox}
\tcbset{
    center title,
    left=0pt,
    right=0pt,
    top=0pt,
    bottom=0pt,
    colframe=black,
    colback=gray!10,
    width=\linewidth,
    enlarge left by=0mm,
    boxsep=5pt,
    boxrule=1pt,
    arc=0pt,outer arc=0pt,
}

\newtcolorbox{promptbox}[1][]{
    enhanced,
    colback=white,
    colframe=black,
    fonttitle=\bfseries,
    title=Prompt,
    attach boxed title to top left={xshift=10pt, yshift*=-\tcboxedtitleheight/2},
    boxed title style={colback=black},
    top=12pt, bottom=10pt, left=10pt, right=10pt,
    #1
}

\tcbset{
    fancybox/.style={
        enhanced,
        breakable,
        arc=1mm,
        outer arc=1mm,
        boxrule=0pt,
        leftrule=1mm,
        toptitle=0.5mm,
        bottomtitle=0.5mm,
        fonttitle=\bfseries\sffamily\small,
        fontupper=\scriptsize,
        coltitle=white,
        drop shadow
    }
}

\newtcolorbox{thoughtbox}{
    fancybox,
    colback=boxyellow,
    colframe=frameyellow,
    coltitle=black,
    title=Thought
}
\newtcolorbox{userbox}{
    fancybox,
    colback=boxpurple,
    colframe=framepurple,
    title=User
}
\newtcolorbox{agentbox}{
    fancybox,
    colback=boxblue,
    colframe=frameblue,
    title=Agent
}
\newtcolorbox{outputbox}{
    fancybox,
    colback=boxgray, 
    colframe=framegray,
    coltitle=black,
    title=Execution Output
}
\newtcolorbox{solutionbox}{
    fancybox,
    colback=boxgreen,
    colframe=framegreen,
    title=Solution
}

\definecolor{codegreen}{rgb}{0.0, 0.5, 0.0}
\definecolor{codegray}{rgb}{0.4, 0.4, 0.4}
\definecolor{codepurple}{rgb}{0.50, 0, 0.50}
\definecolor{backcolour}{rgb}{0.97, 0.97, 0.97}

\lstdefinestyle{mystyle}{
    backgroundcolor=\color{backcolour},
    commentstyle=\color{codegreen},
    keywordstyle=\color{magenta},
    stringstyle=\color{codepurple},
    basicstyle=\ttfamily\scriptsize, 
    breakatwhitespace=false,
    breaklines=true,
    captionpos=b,
    keepspaces=true,
    numbers=none,              
    showspaces=false,
    showstringspaces=false,
    showtabs=false,
    tabsize=2,
    frame=single,
    rulecolor=\color{black!10}, 
    frameround=fttt,            
    upquote=true
}
\title{
\ours: An Open Exploration of Wearable Motion Foundation Models
}

\author[1$*$]{Zongzhe Xu}
\author[1$*$]{Aakarsh Anand}
\author[2$*$]{Sarah Jiang}
\author[3]{Chuntung Zhuang}
\author[1]{Zitao Shuai}
\author[1]{Sriram Sankararaman}
\author[1$\dagger$]{Yuzhe Yang}

\affil[1]{University of California, Los Angeles}
\affil[2]{Duke University}
\affil[3]{John Hopkins University}

\correspondingauthor{$^*$Equal contribution. $^\dagger$Correspondence to: yuzhey@ucla.edu.}

\codelink{https://github.com/yang-ai-lab/Inertia-1}
\projecturl{https://yang-ai-lab.github.io/Inertia-1}

\input{sections/0_abstract}

\begin{document}

\maketitle

\newenvironment{Itemize}{
    \begin{itemize}[leftmargin=*]
    \setlength{\itemsep}{0pt}
    \setlength{\topsep}{0pt}
    \setlength{\partopsep}{0pt}
    \setlength{\parskip}{1pt}}
{\end{itemize}}
\setlength{\leftmargini}{9pt}

\input{sections/0_abstract}
\input{sections/1_intro}
\input{sections/2_related_work}
\input{sections/3_methods}
\input{sections/4_results}
\input{sections/5_discussion}

\section*{Acknowledgments}
We gratefully acknowledge the support by the Amazon Science Hub, the NVIDIA Academic Grant Program, and UCLA DataX. Any opinions, findings, conclusions, or recommendations expressed in this material are those of the author(s) and do not necessarily reflect the views of the funders.

\bibliography{ref}
\bibliographystyle{plain}

\newpage
\appendix
\input{sections/appendix}

\end{document}

%% file: math_commands.tex
\usepackage{amsmath,amsfonts,bm}

\def\figref#1{figure~\ref{#1}}

\def\eqref#1{equation~\ref{#1}}

\def\1{\bm{1}}

\DeclareMathAlphabet{\mathsfit}{\encodingdefault}{\sfdefault}{m}{sl}
\SetMathAlphabet{\mathsfit}{bold}{\encodingdefault}{\sfdefault}{bx}{n}

%% file: sections/0_abstract.tex
\begin{abstract}
\vspace{-7pt}
Wearable motion sensing provides a continuous and scalable window into human behavior and health, making it a natural fit for foundation models, yet its pretraining and scaling principles remain poorly understood. Prior work studies isolated design choices, such as sensor placement or sampling frequency, often under fixed settings and narrow downstream tasks that fail to capture real-world sensing diversity.
We introduce \ours, a fully open exploration of wearable motion foundation models.
Using massive corpora of accelerometer data from global sources spanning more than 18.2M hours, we build a controlled framework for studying the {\textit{full lifecycle}} of wearable motion foundation models, covering \textit{data choices} such as sensor modality, device placement, sampling rate, window length; \textit{model choices} such as architectures and model size; and \textit{training choices} such as pretraining objective and data scale.
Extensive evaluations across 15 datasets spanning \textit{human activity recognition}, \textit{freezing-of-gait detection}, and \textit{disease prediction} reveal intriguing findings for building motion foundation models that generalize across tasks and sensing conditions.
Collectively, \ours not only presents state-of-the-art recipes for diverse downstream tasks, but also serves as a comprehensive, practical, and open cookbook for wearable motion representation learning.

\end{abstract}

%% file: sections/1_intro.tex
\section{Introduction}
\label{sec:intro}

Foundation models turn scale into reusable structure \cite{xu2026sleeplm, shuai2026osf, zhang2025sensorlm}. For wearable intelligence, \textit{motion} is one of the most natural sources of such structure: continuous accelerometer and gyroscope streams record how people move, rest, and function in daily life, linking everyday behavior to mobility \cite{daphnet_freezing_of_gait_245, chan2024capture, li2026hearts}, physiology \cite{narayanswamy2025scaling, yang2022artificial}, and personal health \cite{metwally2026insulin, doherty2017large}.
Large population cohorts now collect longitudinal motion recordings over days or weeks \cite{doherty2017large, chen2011china, patten2026all}, supporting applications from daily activity analysis and mobility assessment to disease risk stratification.
Major studies such as UK Biobank \cite{doherty2017large} contain \textit{millions of hours} of motion signals, yet most of these data remain unlabeled and underused. This creates a clear potential for wearable motion foundation models: learning general-purpose models from large unlabeled cohorts and transferring them across diverse behavioral and health tasks.

Realizing this potential, however, is hindered by a \textit{fragmented} research landscape.
Such fragmentation is layered: \textit{\textbf{data pipelines}} differ in sampling rate, window length, sensor modality, body placement, and axis representation; \textit{\textbf{methodologies}} differ in pretraining objective, architecture, representation domain, and training recipe; \textbf{\textit{datasets}} differ in scale, duration, label density, and clinical context; and \textit{\textbf{downstream evaluations}} differ across human activity recognition (\har), gait analysis, and disease prediction. As summarized in Table \ref{tab:wearables-related-work}, existing motion studies typically cover only a subset of sensors, placements, tasks, and scales, leaving the field with strong local findings but limited guidance for building a general-purpose wearable motion foundation model.

\begin{figure}[!t]
\centering
\includegraphics[width=\textwidth]{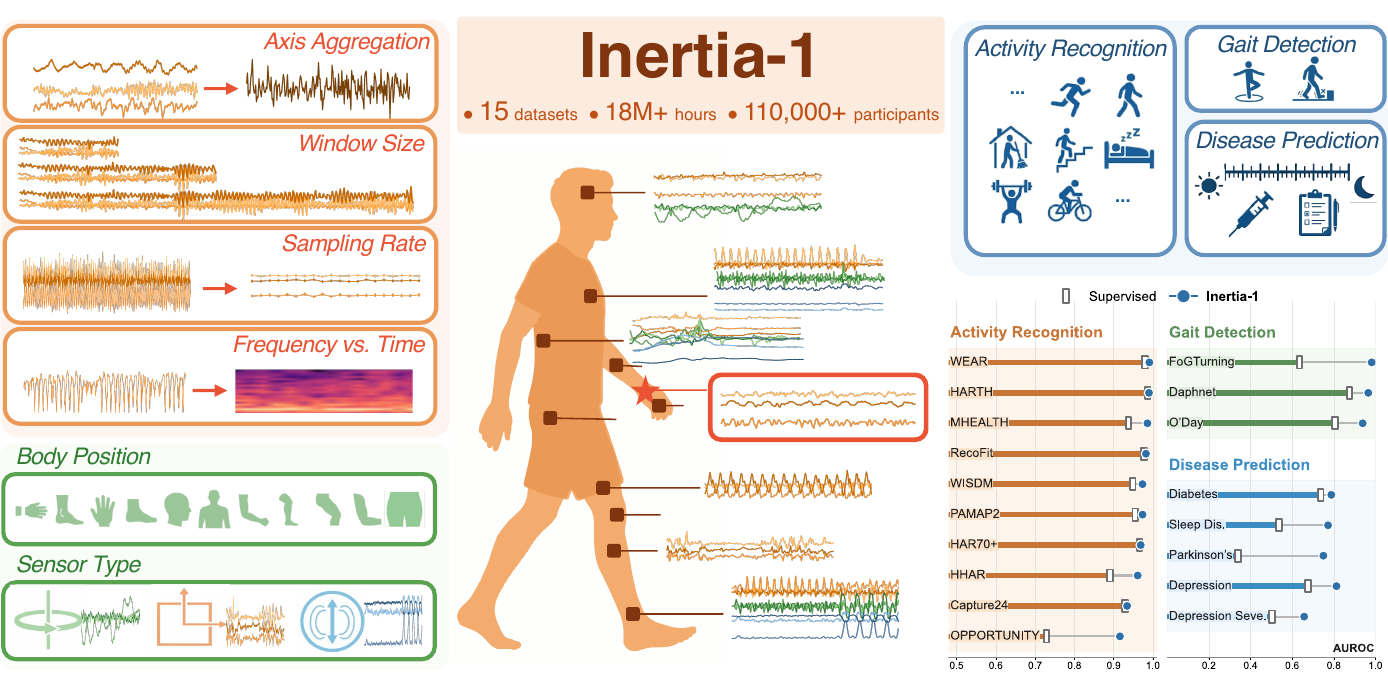}
\caption{\small{
\textbf{Overview of \ours.}
We present a unified and fully open exploration of wearable motion FMs, spanning large-scale accelerometer pretraining, diverse self-supervised objectives, controlled sensing setups, and downstream evaluation across activity recognition, gait analysis, and disease prediction. \ours covers 18.2M hours of motion data from over 115,000 individuals and 15 datasets, enabling systematic study of how data, model, and training choices shape transferable motion representations. More details are in Appendix \ref{app:data_eval}.
}}
\label{fig:overview}
\vspace{-12pt}
\end{figure}

Specifically, the nature of this fragmentation is twofold.
\textbf{First}, \textit{data configurations} and \textit{modeling methods} are rarely examined under a shared protocol. Prior work has studied diverse but isolated design factors such as windowing \cite{banos2014window}, sensor displacement \cite{banos2014displacement}, and device heterogeneity \cite{stisen2015smart} across disjoint datasets, while recent foundation-model efforts explore reconstructive \cite{miao2024spatial}, contrastive \cite{yang2023simper}, self-distillation \cite{vu2025smooth}, and frequency-domain objectives \cite{logacjov2024selfpab}, often under fixed input settings. Yet, when objectives, preprocessing choices, sensing setups, and representation domains all vary together, it becomes difficult to attribute progress to \textit{any} single design choice.
\textbf{Second}, \textit{dataset scale} and \textit{task definition} are split across regimes. \textit{Large-scale} cohorts provide population-level longitudinal data and health outcomes, but often release compressed motion representations, such as downsampled traces or vector-magnitude summaries \cite{nebeker2026sharing, patten2026all}. In contrast, datasets for \har and freezing-of-gait (\fog) offer fine-grained, high-frequency signals, but are inherently \textit{small-scale}, with limited participants and short sessions \cite{daphnet_freezing_of_gait_245}. 
This leaves open whether massive unlabeled cohorts can produce representations that bridge both small, label-scarce motion tasks and population-level disease modeling.

To fill the gaps, we present \ours, an open exploration of wearable motion foundation models. Rather than simply proposing a new architecture, \ours is designed to disentangle how pretraining objectives, sensing configurations, scale, and downstream tasks jointly shape transferable motion representations (see \figref{fig:overview}):
\ding{182} \textbf{Pretraining.} We integrate {10} representative methods under the same training and evaluation settings, covering {5} classes of pretraining objectives, including general self-supervised as well as motion-specific approaches.
\ding{183} \textbf{Data pipelines.} We evaluate models across a rigorous grid of experimental axes, including sampling rate, window length, sensor modality, sensor axis dimensionality, and body placement (\tabref{tab:inertia1_breadth}).
\ding{184} \textbf{Scale.} We aggregate, to our knowledge, the largest waveform-level wearable motion data collection to date, spanning {18.2M} hours of motion data from more than {115,000} people and {15} datasets (\tabref{tab:wearables-related-work}).
\ding{185} \textbf{Downstream tasks.} We evaluate across {10} human activity recognition datasets, {3} freezing-of-gait datasets, and {7} disease prediction tasks, spanning both fine-grained motion understanding and population-level health modeling.

\ours is designed to be easily extensible, supporting new models, datasets, and tasks as the field evolves.
With the unified framework and more than {1,000} trained models, we reveal intriguing lessons and  future directions for building state-of-the-art wearable motion foundation models:
\vspace{-6pt}
\begin{Itemize}
\item \textit{Pretraining is beneficial, but not universal --} Self-supervised pretraining consistently improves over supervised counterparts, yet no single objective dominates across all task families and metrics.
\vspace{2pt}
\item \textit{Data representation is a first-order modeling choice --} Triaxial motion signals outperform magnitude summaries, while sampling rate and window length matter differently across task granularities.
\vspace{2pt}
\item \textit{Scale helps, but not all scale is equal --} More (diverse) pretraining data yields steadier gains than larger model sizes, and multi-sensor fusion lead to strong gains in sensor-expansion settings.
\vspace{2pt}
\item \textit{State-of-the-art motion foundation models require coordinated design --} Performance depends on the joint choice of data fidelity, sensing setup, objective, and scale, rather than any single recipe.
\vspace{2pt}
\item \textit{Inertia-1 as an open cookbook --} As to date the most diverse and unified testbed, \ours offers a practical foundation for studying, extending, and deploying wearable motion foundation models.
\end{Itemize}

%% file: sections/2_related_work.tex
\vspace{-1pt}
\section{Related Work}
\label{sec:related-work}
\vspace{-5pt}

\input{tables/main_related_work}

\textbf{Applications of Wearable Motion Modeling.}
Wearable motion signals support tasks across very different temporal and clinical scales. \har classifies short windows of inertial data into activities such as walking, sitting, running, or household actions \cite{9257355, 6365160}, while \fog detection identifies brief clinically meaningful gait events in Parkinson's disease \cite{10.3389/fnagi.2023.1119956,rodriguezmartin2017fog}. At a larger scale, population studies use multi-day accelerometry to quantify rest-activity rhythms, sleep, mobility, sedentary behavior, and disease risk \cite{Master_Annis_Huang_Beckman_Ratsimbazafy_Marginean_Carroll_Natarajan_Harrell_Roden_et_al_2022,Shim2023,doherty2017large}. These areas have usually evolved separately: \har emphasizes fine-grained labels and high-resolution windows, \fog emphasizes symptom-specific detection in targeted cohorts, and population-health studies emphasize aggregate longitudinal behavior. In contrast, \ours studies transfer across these regimes under \textit{one unified} framework, with to date the broadest coverage, including ten \har datasets, three \fog datasets, and six disease prediction tasks (\tabref{tab:wearables-related-work}).

\textbf{Sensing Configurations and Deployment Settings.}
Wearable motion models are shaped not only by architecture, but also by how signals are collected and represented. Prior work has shown that window length affects recognition accuracy and latency \cite{banos2014window}, body placement and displacement can cause large performance shifts \cite{banos2014displacement,info:doi/10.2196/23681}, and device heterogeneity leads to variation in sensing behavior, hardware, and operating-system processing \cite{stisen2015smart}. Sampling rate, axis representation, and sensor modality also determine how much temporal and spatial motion structure is available \cite{yamane2025effects, huang2022smartphone}. These factors are usually studied isolated at a time, often in supervised \har settings and outside modern large-scale pretraining. \ours revisits these sensing and deployment choices under a unified pretraining setup, testing sampling frequency, window length, triaxial \textit{vs.} reduced-axis input, time- \textit{vs.} frequency-domain representation, sensor modality, placement, model size, and data scale (\tabref{tab:wearables-related-work}).

\textbf{Self-Supervised Learning and Motion Foundation Models.}
Self-supervised learning (SSL) is a natural fit for wearable motion because unlabeled accelerometry is abundant, while dense labels are costly and task-specific. Early wearable SSL studies explored pretext and contrastive objectives for \har \cite{haresamudram2022ssl,tang2021selfhar}, while recent methods adapt masked reconstruction, contrastive, temporal prediction, self-distillation, and frequency-domain modeling to inertial signals \cite{yuan2024self, logacjov2024selfpab, yang2023simper, 10.1145/3410531.3414306}. A parallel line scales pretraining to large cohorts such as UK Biobank \cite{doherty2017large}, showing that large accelerometer corpora can improve activity recognition and health prediction \cite{doherty2017large, nhanes2011, yuan2024self, xu2025lsm2learningincompletewearable}. Related biosignal foundation models (FMs) further show the promise of large-scale physiological pretraining \cite{abbaspourazad2024largescaletrainingfoundationmodels, doi:10.1056/AIoa2401033}. However, existing motion FM studies often focus on a single objective, input format, corpus, or task setting. In contrast, \ours compares representative objectives under controlled conditions and studies how they transfer across activity recognition, gait analysis, and disease prediction, providing practical guidance for building wearable motion FMs that are robust, generalizable, and useful across datasets, tasks, input modalities, and deployment settings.

%% file: tables/main_related_work.tex
\begin{table*}[!t]
\setlength{\tabcolsep}{5pt}
\renewcommand{\arraystretch}{1.1}
\centering
\caption{\small{
\textbf{Comparisons of studies on wearable motion data modeling.} We compare a subset of major device placements; the full set of covered placements in \ours is provided in Table \ref{tab:inertia1_breadth}.}
}
\vspace{-2pt}
\label{tab:wearables-related-work}
\small
\adjustbox{max width=\textwidth}{
\begin{tabular}{lrrrcccccccccccc}
\toprule[1.5pt]
\multirow{2.5}{*}{\textbf{Study}$^\dagger$}
 & \multirow{2.5}{*}{\textbf{\# People}}
 & \multirow{2}{*}{\textbf{\# Hours}}
 & \multirow{2.5}{*}{\textbf{Datasets}}
 & \multicolumn{3}{c}{\textbf{Sensors}}
 & \multicolumn{6}{c}{\textbf{Placements}}
 & \multicolumn{3}{c}{\textbf{Tasks}} \\
\cmidrule(lr){5-7} \cmidrule(lr){8-13} \cmidrule(lr){14-16}
 & & (000s) & 
 & \bACC & \bGYR & \bMAG
 & \bWR & \bAR & \bCH & \bHP & \bLG & \bAN
 & \texttt{HAR} & \texttt{FoG} & \texttt{DP} \\
\midrule\midrule
Banos \textit{et al.} \cite{banos2014displacement} & 17 & $\ll$1 & 1 & \ryes & \ryes & \ryes & \ryes & \ryes & \ryes & \ryes & \ryes & \ryes & \ryes & \rno & \rno \\
Stisen \textit{et al.} \cite{stisen2015smart} & 9 & $\ll$1 & 1 & \ryes & \ryes & \rno & \ryes & \rno & \rno & \ryes & \rno & \rno & \ryes & \rno & \rno \\
Haresamudram \textit{et al.} \cite{haresamudram2022ssl} & 320 & 4 & 10 & \ryes & \rno & \rno & \ryes & \rno & \rno & \ryes & \ryes & \ryes & \ryes & \ryes & \rno \\
Yuan \textit{et al.} \cite{yuan2024self} & 100,000 & 15,700 & 9 & \ryes & \rno & \rno & \ryes & \rno & \rno & \rno & \rno & \rno & \ryes & \rno & \rno \\
Logacjov \textit{et al.} \cite{logacjov2024selfpab} & 35,000 & 100 & 6 & \ryes & \rno & \rno & \ryes & \rno & \ryes & \ryes & \ryes & \ryes & \ryes & \rno & \rno \\
Hoddes \textit{et al.} \cite{hoddes2025scaling} & 60 & 1.6 & 4 & \ryes & \ryes & \rno & \ryes & \rno & \rno & \rno & \rno & \rno & \ryes & \rno & \rno \\
Xu \textit{et al.} \cite{xu2025relcon} & 87,380 & 720 & 7 & \ryes & \rno & \rno & \ryes & \rno & \rno & \ryes & \ryes & \rno & \ryes & \ryes & \rno \\
\midrule
\grayrow
\textbf{\ours (Ours)} & \textbf{115,450} & \textbf{18,224} & \textbf{15} & \ryes & \ryes & \ryes & \ryes & \ryes & \ryes & \ryes & \ryes & \ryes & \ryes & \ryes & \ryes \\
\bottomrule[1.5pt]
\end{tabular}
}
\vspace{4pt}
\newline
\scriptsize
{\setlength{\fboxsep}{1.5pt}%
\bACC: Accelerometer.~~ \bGYR: Gyroscope.~~ \bMAG: Magnetometer.~~~\bWR: Wrist.~~ \bAR: Arm.~~ \bCH: Chest.~~ \bHP: Hip.~~ \bLG: Leg.~~ \bAN: Ankle.}
\vspace{3pt}
\newline
\texttt{HAR}: Human Activity Recognition.~~ \texttt{FoG}: Freezing of Gait Detection.~~ \texttt{DP}: Disease Prediction.~~~ $^\dagger$\,Only raw-waveform studies considered.
\vspace{-5pt}
\end{table*}

%% file: sections/3_methods.tex
\vspace{-5pt}
\section{\emph{Inertia}-1}
\label{sec:method}
\vspace{-5pt}

\begin{wrapfigure}[15]{r}{0.41\textwidth}
\centering
\vspace{-42pt}
\includegraphics[width=0.4\textwidth]{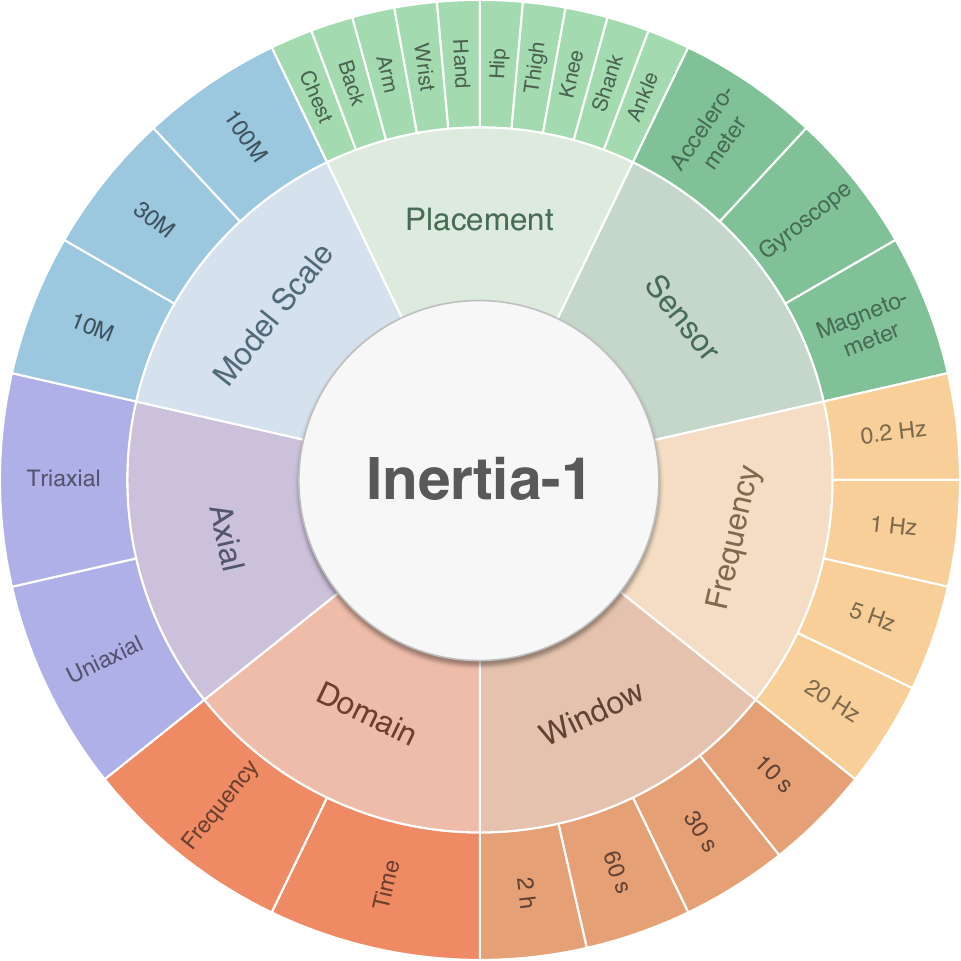}
\caption{\small{
\textbf{Coverage breadth of \ours.} Full settings are included in \tabref{tab:inertia1_breadth}.
}}
\vspace{42pt}
\end{wrapfigure}
\label{fig:inertia1-breadth-circle}

The goal of \ours is to turn the fragmented landscape of wearable motion modeling into a unified space of controlled exploration. Rather than searching for one best architecture, we study motion FMs a joint problem over data, sensing configurations, pretraining algorithms, and downstream tasks. This is essential for motion sensing: the same behavior can appear differently across placements, sampling rates, sensor modalities, and temporal windows, while the same pretrained representation should transfer from short-term activity recognition to long-term disease prediction.

\ours is designed as an open, extensible framework for studying these interactions. It aggregates \textbf{15} datasets across \textbf{3} task families and \textbf{18} downstream tasks, pretrains a broad set of representative objectives, and evaluates models across a controlled grid of sensing and scaling choices. Together, these components allow us to ask not only which model performs best, but which design choices make wearable motion FMs robust, transferable, and useful across real-world conditions. \figref{fig:inertia1-breadth-circle} and \tabref{tab:main:data} summarize the breadth of the explored space.

\vspace{-5pt}

\subsection{Data Regimes and Tasks}
\vspace{-2pt}
A central design goal of \ours is to bridge motion sensing regimes that have been studied separately: large cohorts provide scale, longitudinal coverage, and health outcomes, while targeted motion datasets provide high-frequency signals and fine-grained labels. We use \nhanesCite as 
\input{tables/main_datasets}
the primary pretraining source because it connects these regimes, providing raw high-frequency accelerometer data from {14,000} participants over multiple days, together with rich health and survey variables. We further incorporate the \ukbCite accelerometer data (more than 100,000 participants) in scaling studies to examine how massive lower-resolution cohort data complement high-resolution waveform pretraining. Our downstream evaluation spans three task families, as summarized in \tabref{tab:main:data}.
More datasets details and experimental setups are provided in Appendix \ref{app:data_eval}.

\vspace{-3pt}
\wrapitem{\textbf{Human activity recognition (\textit{short-term}):} We include \textbf{10} HAR datasets spanning daily routines, free-living behavior, laboratory activities, and fine-grained exercise movements across diverse environments and sensor setups. These tasks evaluate whether pretrained representations capture short-term motion semantics such as walking, sitting, running, and household activity.}
\vspace{4pt}
\wrapitem{\textbf{Freezing-of-gait detection (\textit{transient}):} We select \textbf{3} \fog datasets targeting brief, clinically meaningful disruptions in locomotion. These tasks test whether motion representations pretrained on broad wearable data can detect subtle, transient abnormalities beyond standard activity categories.}
\vspace{4pt}
\wrapitem{\textbf{Disease prediction (\textit{long-term}):} We include \textbf{7} disease-related tasks from \nhanes and \ukb, covering population-level health modeling from longitudinal motion patterns. Unlike \har and \fog, these tasks are not tied to isolated short-window actions, but to broader behavioral signatures.} 

\vspace{-6pt}
\subsection{Pretraining Algorithms}
\vspace{-3pt}
To study how different learning principles interact with wearable motion data, \ours compares representative pretraining algorithms under shared data, sensing, and evaluation settings. We include both (1) \textit{general SSL} objectives, and (2) \textit{wearable-specific SSL} methods, and compare them with \textit{supervised baselines} trained from scratch on each downstream task. In total, we cover \textbf{10} (pre)training approaches that represent the state-of-the-art in wearable motion FM. This design lets us separate whether pretraining helps from which form of pretraining is most useful for transferable motion representations. Additional details for each model are provided in Appendix \ref{app:pretraining_algorithms}.

\vspace{-5pt}
\begin{Itemize}
\item \textbf{Autoregressive prediction.} GRU-based (\argruCite) and Transformer-based (\artransformerCite) models learn representations by predicting future motion from past context, testing whether sequential forecasting objectives capture transferable behavioral dynamics.
\vspace{4pt}
\item \textbf{Contrastive learning.} \simclrCite, \sslwearablesCite, and \relconCite learn by aligning related motion views while separating unrelated ones. They cover both general contrastive learning and wearable-specific objectives designed around motion augmentations and temporal structure.
\vspace{4pt}
\item \textbf{Self-distillation.} \dinoCite learns stable representations through teacher-student consistency training, testing whether non-contrastive representation learning transfers to wearable motion.

\vspace{4pt}
\item \textbf{Masked reconstruction.} \patchtstCite, \selfpabCite, and \lsmCite learn by recovering missing signal segments, covering both time-domain and time-frequency reconstruction-style objectives.
\end{Itemize}

\subsection{Axes of Exploration}
\vspace{-4pt}

\textbf{Disentangling Sensing Configurations.} To understand how motion FMs interact with with the physical constraints of wearable data, we evaluate selected pretraining objectives across four core input axes below: 

\wrapitem{\textit{Uniaxial vs. Triaxial Input:} Large population cohorts often release vector-magnitude summaries of acceleration to reduce data footprint and privacy risk \cite{doherty2017large, patten2026all}, whereas targeted motion datasets usually provide full triaxial signals \cite{chan2024capture, opportunity_activity_recognition_226}. We compare these settings to quantify information loss from axis aggregation.}
\vspace{-2pt}
\wrapitem{\textit{Window Length:} Prior work often favors short windows to isolate atomic movements and balance accuracy with latency \cite{banos2014window}. $\{10s, 30s, 60s, 2h\}$ windows are used to test how temporal context affects diverse tasks across \har, \fog, and disease prediction.}
\vspace{-2pt}
\wrapitem{\textit{Sampling Frequency:} Motion datasets vary widely in sampling rate, from high-frequency segments to low-resolution monitoring. We downsample raw accelerometer to $\{20Hz, 5Hz, 1Hz,  0.2Hz\}$ to test when temporal resolution is essential and whether pretraining can offset downsampling \cite{yamane2025effects}.}
\vspace{-2pt}
\wrapitem{\textit{Representation Domain:} Time-domain models are widely used for modern wearable SSL \cite{yuan2024self, tang2021selfhar}, while frequency-domain methods may improve invariance to noise and phase shifts \cite{logacjov2024selfpab, yang2023simper}. We compare them under matched settings to answer which domain transfers better across tasks.}

\textbf{Spatial and Modality Robustness.} Large-scale pretraining datasets are often dominated by wrist-worn accelerometers, whereas real-world and clinical studies involve more diverse placements and sensor suites. We therefore evaluate whether representations learned from common consumer-style sensing transfer to heterogeneous body locations and additional inertial modalities.
\vspace{-5pt}
\begin{Itemize}
    \item \textit{Body Placements:} Prior work often treats placement shift as a domain adaptation problem \cite{chakma2023domain, an2021adaptnet, sanabria2021unsupervised}. In contrast, we study direct transfer to other placements after large-scale pretraining on wrist-worn accelerometers, and test whether multi-placement fusion improves over single-placement ones.
    \vspace{3pt}
    \item \textit{Sensor Modalities:} Most cohort-scale data are accelerometer-based, whereas many high-fidelity datasets also include gyroscopes and magnetometers. We test whether accelerometer-pretrained models transfer to these modalities and whether additional sensors provide complementary benefits.
\end{Itemize}
\vspace{-3pt}

\textbf{Model and Data Scaling.} 
Finally, we examine whether scale can help resolve fragmented design choices in wearable motion modeling. We vary model capacity from small ($\sim$5M) to medium ($\sim$30M) and large ($\sim$100M), and scale pretraining data volume using \nhanesCite and \ukbCite. 

%% file: tables/main_datasets.tex
\begin{wraptable}[19]{r}{0.5\textwidth}
\vspace{-8pt}
\setlength{\tabcolsep}{3pt}
\renewcommand{\arraystretch}{1.05}
\centering
\caption{\small{
\textbf{Overview of \ours datasets.} Detailed statistics and setups are in Appendix \ref{app:data_eval} and Table \ref{tab:appendix:data}.
}}
\vspace{-5pt}
{\small
\adjustbox{max width=0.5\textwidth}{
\begin{tabular}{llll}
\toprule[1.5pt]
\textbf{Dataset} & \textbf{Task} & \textbf{Sensor} & \textbf{Placement} \\
\midrule
\captureCite  & \har & \bACC & \bWR \\
\harplusCite & \har & \bACC & \bCH \bLG \\
\harthCite & \har & \bACC & \bCH \bLG \\
\hharCite & \har & \bACC \bGYR & \bWR \\
\mhealthCite & \har & \bACC \bGYR \bMAG & \bAR \bCH \bAN \\
\opportunityCite & \har & \bACC \bGYR \bMAG & \bWR \bHN \bAR \\
&&& \bBK \bHP \bKN \\
\pamapCite & \har & \bACC \bGYR \bMAG & \bHN \bCH \bAN \\
\recofitCite & \har & \bACC \bGYR & \bAR \\
\wisdmCite & \har & \bACC \bGYR & \bWR \\
\wearCite & \har & \bACC & \bAR \bLG \\
\daphnetCite & \fog & \bACC & \bHP \bLG \bAN \\
\odayfogCite & \fog & \bACC \bGYR & \bWR \bCH \bHP \\
&&& \bLG \bAN \\
\fogturningCite & \fog & \bACC \bGYR & \bSH \\
\nhanesCite & {Disease} & \bACC & \bWR \\
\ukbCite & {Disease} & \bACC & \bWR \\
\bottomrule[1.5pt]
\end{tabular}
}}
\end{wraptable}
\label{tab:main:data}

%% file: sections/4_results.tex
\vspace{-2pt}
\section{Results and Analyses}
\vspace{-4pt}

\subsection{Identifying Robust Pretraining Objectives}
\label{subsec:model_compare}
\vspace{-4pt}

We first compare our suite of 10 model objectives against supervised baselines to identify which methods remain robust across the full task spectrum. Under the default setting of 20Hz sampling, 30s windows, and triaxial inputs, all models are pretrained on \nhanes. We evaluate transfer through both full finetuning and linear probing across all \har and \fog tasks. For disease prediction, we use frozen pretrained backbones with a specialized multi-instance learning (MIL) head to produce daily-level predictions \cite{shuai2026osf}. Supervised baselines are trained from scratch on each downstream task. Additional training and evaluation details are provided in Appendix \ref{app:data_eval} \& \ref{app:pretraining_algorithms}.

\input{tables/main-model-comp}

\input{tables/main-model-fog}

\textbf{Large-scale pretraining improves transfer across task families.}
As shown in Table \ref{tab:har_auroc}, \ref{tab:fog_auroc}, and \ref{tab:disease_auroc}, self-supervised pretraining consistently outperforms supervised training from scratch across task categories. This suggests strong representations learned from unlabeled population cohorts \textit{regardless of the specific downstream application}. While no single objective dominates all settings, specialized or predictive objectives are generally more robust for short-term recognition tasks than generic augmentation-based baselines, while these differences are less pronounced in disease prediction. 
\begin{wraptable}{r}{0.45\textwidth}
\centering
\small
\vspace{0pt}
\caption{\small{\textbf{Task variation in model performance.}
\textit{SSL variance} measures the task-level average gap between the best and median SSL AUROC. 
\textit{SSL gain} measures the average gap between the best SSL AUROC and the best supervised AUROC.}}
\label{tab:task-performance-gap}
\setlength{\tabcolsep}{3pt}
\begin{adjustbox}{width=0.95\linewidth}
\begin{tabular}{lcc}
\toprule[1.5pt]
Task Type & SSL Variance & SSL Gains \\
\midrule
\midrule
\har & \graycell{\textbf{2.5}} & 4.0 \\
\fog & 11.0 & \graycell{\textbf{16.4}} \\
Disease Prediction & 10.0 & \graycell{\textbf{19.7}} \\
\bottomrule[1.5pt]
\end{tabular}
\end{adjustbox}
\vspace{-20pt}
\end{wraptable}
Overall, objective design remains important, but its benefit depends on the temporal structure and semantic granularity of the downstream task rather than a universally optimal pretraining strategy.

\textbf{Task variation widens SSL variances and gains.} 
While differences between SSL objectives are modest for standard \har, they grow significantly for other specialized tasks. Table \ref{tab:task-performance-gap} measures this using \textit{\textbf{SSL variance}}, defined as the gap between the best and median SSL AUROC. This value is only 2.5 for \har, but increases to 11.0 for \fog and 10.0 for disease prediction. \textit{\textbf{SSL gains}} over supervised training follow the same pattern, rising from 4.0 on \har to 16.4 on \fog and 19.7 on disease prediction. These results suggest that basic HAR is closer to saturation, whereas clinical and disease-oriented tasks still benefit substantially from representations that capture subtle motion and physiological signatures.

\subsection{When Does Scaling Help Motion Foundation Models?}

\begin{wrapfigure}{r}{0.55\textwidth}
\vspace{-10pt}
\centering
\begin{minipage}{\linewidth}
\centering
\includegraphics[width=\linewidth]{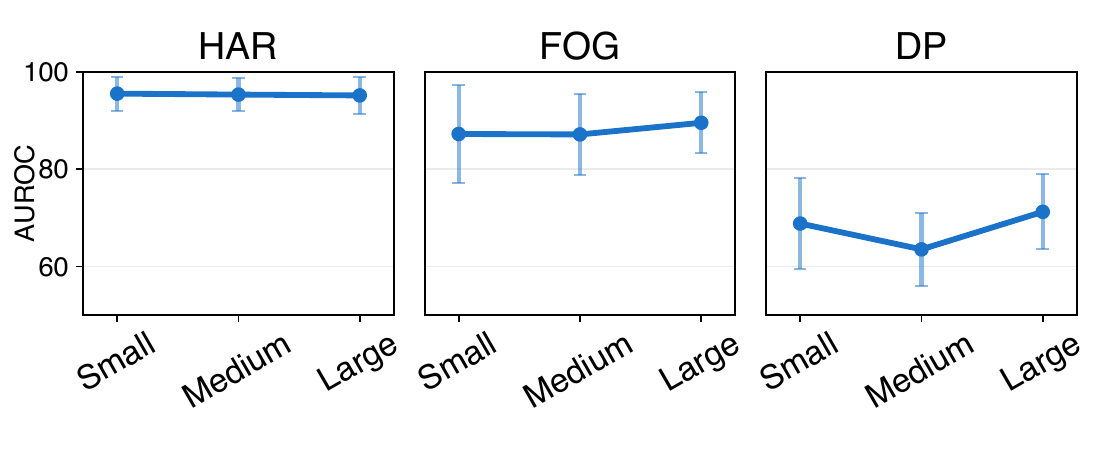}
\vspace{-25pt}
\captionof{figure}{\small{\textbf{\textcolor{modelscaling}{Model scaling} across tasks.} We vary the model size and report linear probe performance across all tasks.}}
\label{fig:artr-scaling}
\vspace{8pt}
\includegraphics[width=\linewidth]{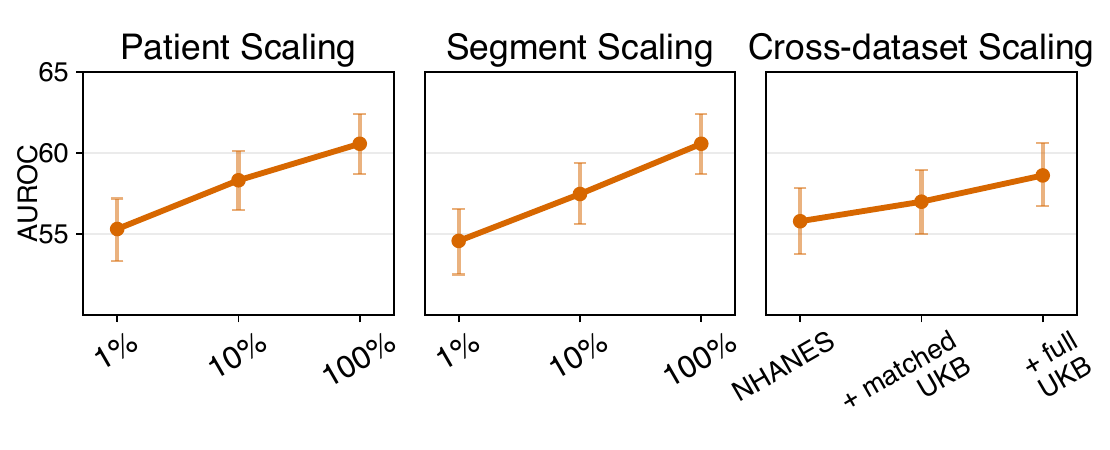}
\vspace{-25pt}
\captionof{figure}{\small{\textbf{\textcolor{datascaling}{Data scaling} across settings.} We compare \ukb disease-prediction performance when scaling \textbf{(left)} the number of individuals, \textbf{(middle)} the number of segments per individual, and \textbf{(right)} cross-dataset data mixtures.}}
\label{fig:scaling}
\end{minipage}
\vspace{-15pt}
\end{wrapfigure}

\textbf{Model capacity saturates under fixed data.}
We first ask whether increasing encoder size alone improves motion representations. Under the default setting, we scale \artransformer from 10M to 30M and 100M parameters while keeping the pretraining corpus and input configuration fixed. As shown in Fig. \ref{fig:artr-scaling}, linear-probe performance largely plateaus across model sizes and task families, suggesting that additional capacity does not translate into better representations without additional data or stronger task alignment. Under full finetuning, larger models can even degrade performance, likely due to overfitting on smaller downstream datasets (Appendix \ref{app:supp_scaling}).
These results suggest that motion FMs are less limited by parameter count than by the diversity and relevance of the pretraining signal.

\textbf{Data scale improves transfer through both population diversity and behavioral coverage.}
We next test whether the scaling bottleneck can be reduced by expanding the pretraining data, using \ukb as a large-cohort testbed and evaluating frozen representations on patient-disjoint disease prediction tasks. As demonstrated in Fig. \ref{fig:scaling} \& \ref{fig:ukb_scaling_all}, increasing both (1) the \textit{number of individuals}, and (2) the \textit{number of segments} per individual improves AUROC, suggesting that wearable disease representations benefit from population diversity as well as richer behavioral coverage within each 
person. We also observe gains when mixing \ukb with \nhanes during pretraining, indicating that data diversity can improve transfer even when the original corpus is already large. 

\textbf{Scaling is data-first, but not model-free.}
Together, these experiments suggest that data volume and diversity provide a more reliable path to improved transfer than increasing model size alone. This does not mean model capacity is unimportant; rather, larger encoders may require richer pretraining data, better objective design, or more careful tuning before their capacity yields consistent downstream gains. In this sense, scaling wearable motion FMs is not only a question of bigger networks, but of matching model capacity to signal diversity, temporal resolution, and downstream task structure. We provide more results and experimental details in Appendix \ref{app:supp_scaling}.\looseness=-1

\subsection{A Unified Lens over Sensing Design Space}

To isolate the effect of core sensing configurations, we conduct controlled experiments using \artransformer and \patchtst, two robust methods from the objective comparison. We use 30s windows, 20Hz sampling, 
\begin{wrapfigure}{r}{0.62\textwidth}
\vspace{-10pt}
\centering

\begin{minipage}[t]{0.48\linewidth}
\centering

\includegraphics[width=0.95\linewidth]{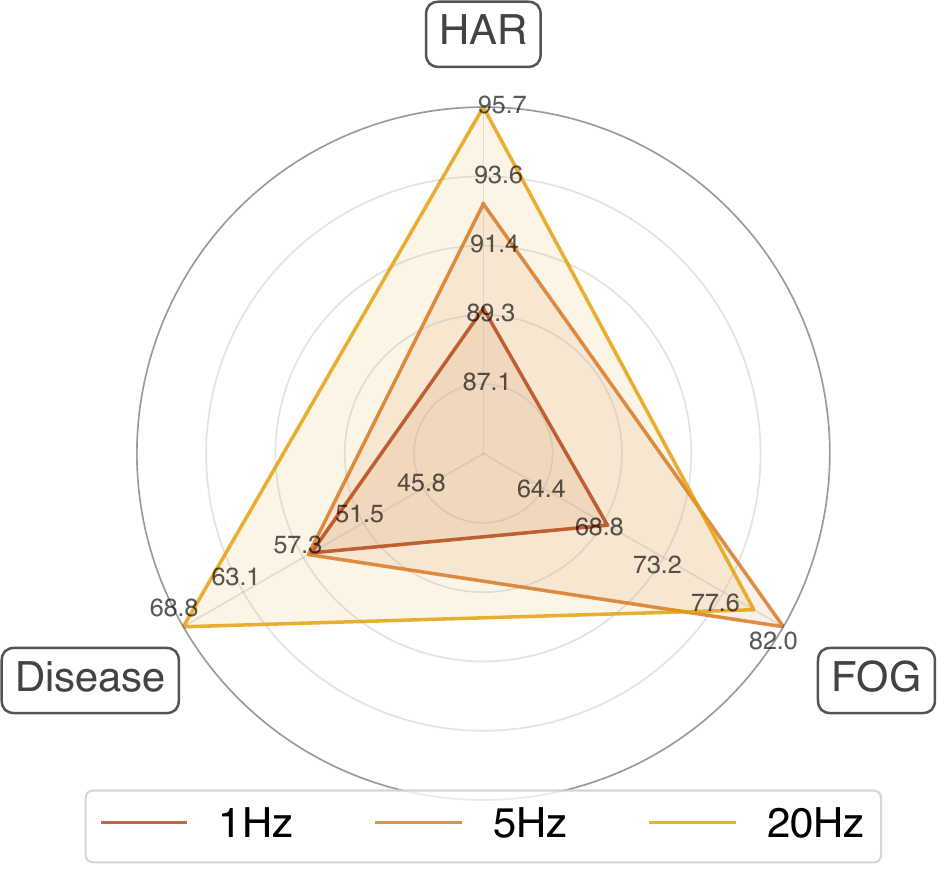}

\vspace{0pt}

\scriptsize
\setlength{\tabcolsep}{2.5pt}
\begin{tabular}{p{1cm}cccc}
\toprule[1pt]
Freq. & HAR & FOG & DP & Overall \\
\midrule
1Hz & 89.5 & 69.1 & 56.5 & 71.7 \\
5Hz & 92.7 & \textbf{82.0} & 54.4 & 76.4 \\
20Hz & \textbf{95.7} & 79.8 & \textbf{68.8} & \textbf{81.5} \\
\bottomrule[1pt]
\end{tabular}

\vspace{2pt}
{\footnotesize\textbf{Sampling frequency}}
\end{minipage}
\hfill
\begin{minipage}[t]{0.48\linewidth}
\centering

\includegraphics[width=0.95\linewidth]{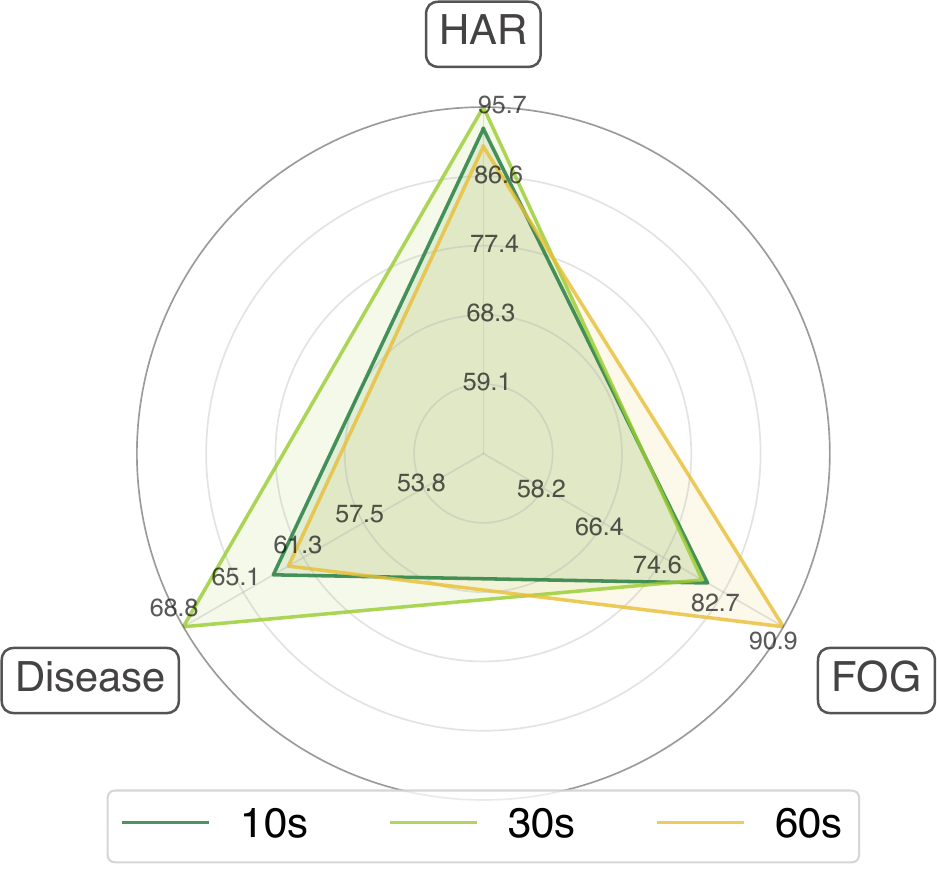}

\vspace{0pt}

\scriptsize
\setlength{\tabcolsep}{2.5pt}
\begin{tabular}{p{1cm}cccc}
\toprule[1pt]
Window & HAR & FoG & DP & Overall \\
\midrule
10s & 92.9 & 80.6 & 63.2 & 78.9 \\
30s & \textbf{95.7} & 79.8 & \textbf{68.8} & \textbf{81.5} \\
60s & 90.5 & \textbf{90.9} & 62.2 & 81.2 \\
\bottomrule[1pt]
\end{tabular}

\vspace{2pt}
{\footnotesize\textbf{Window size}}
\end{minipage}

\caption{\small{
\textbf{Control studies: sampling frequencies and window sizes.} \textbf{(Left)} Full-finetuning AUROC across sampling rates. \textbf{(Right)} Full-finetuning AUROC across window sizes. Lower sampling rates remain competitive for \har, while disease prediction benefits from higher temporal resolution; window length effects are task-dependent.}}
\label{fig:temporal-ablation}
\vspace{-10pt}
\end{wrapfigure}
and triaxial accelerometer inputs as the default setting, then vary one axis at a time, including sampling rate, window length, axis dimensionality, and representation domain. For each setting, we pretrain a new model on the same corpus and finetune it across downstream tasks following the protocol in Sec. \ref{subsec:model_compare}. We report \artransformer results in Fig. \ref{fig:temporal-ablation} and Fig. \ref{fig:input-ablation}; \patchtst results and additional details are provided in Appendix \ref{app:controlled_config}.

\textbf{Pretraining improves robustness to low-frequency sampling.}
As shown in Fig. \ref{fig:temporal-ablation}, performance generally improves with sampling frequency, but the degradation at lower rates is smaller than expected for \har. Unlike prior studies that report sharp drops below 10Hz \cite{yamane2025effects}, pretrained representations remain competitive even at 1Hz, suggesting that large-scale pretraining captures contextual motion patterns that partly compensate for lost high-frequency detail. This robustness is weaker for disease prediction, where higher sampling rates provide clearer gains, indicating that clinical outcomes may depend on subtle biomechanical signatures that are attenuated by aggressive downsampling.

\textbf{Window length is task-dependent.}
Window size has a more mixed effect. Shorter windows preserve discrete movement semantics and reduce label mixing, while longer windows provide broader temporal context. As shown in Fig. \ref{fig:temporal-ablation}, 30s and 60s windows perform strongly overall, but no single duration is optimal across \har, \fog, and \dd. This suggests that temporal context
\begin{wrapfigure}{r}{0.62\textwidth}
\vspace{-8pt}
\centering

\begin{minipage}[t]{0.48\linewidth}
\centering
\includegraphics[width=0.95\linewidth]{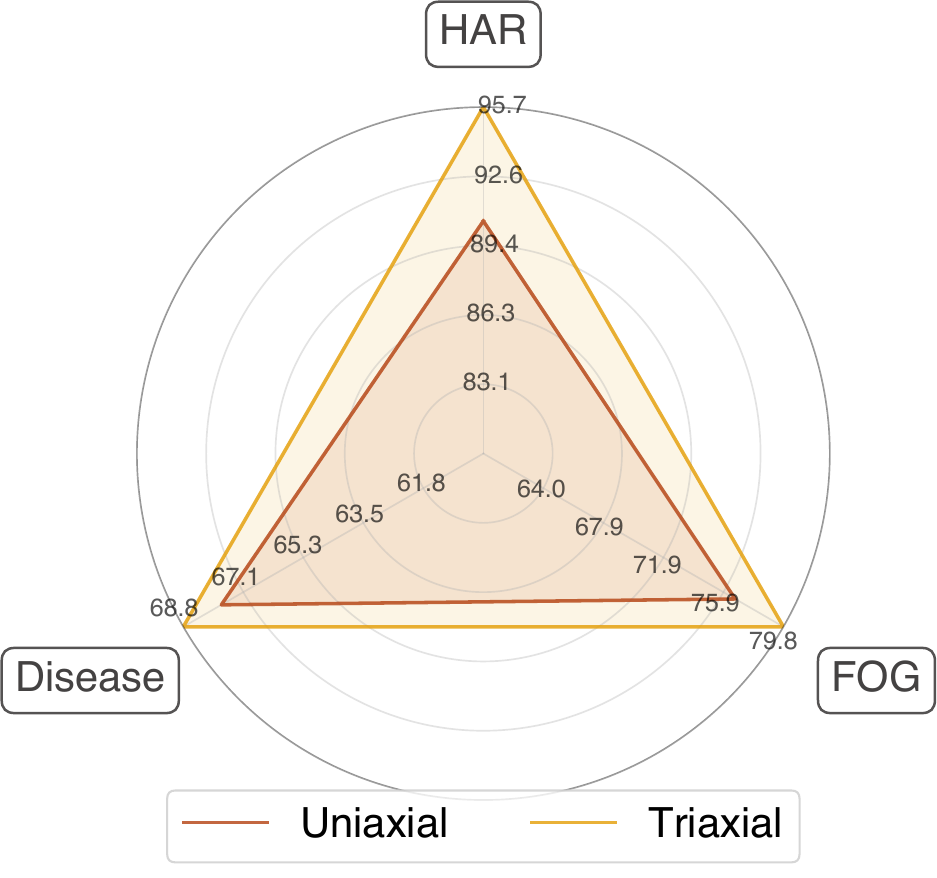}

\vspace{0pt}
\scriptsize
\setlength{\tabcolsep}{2.5pt}
\begin{tabular}{p{1cm}cccc}
\toprule[1pt]
Axes & HAR & FoG & DP & Overall \\
\midrule
Uniaxial & 90.6 & 76.7 & 67.7 & 78.3 \\
\grayrow
Triaxial & \textbf{95.7} & \textbf{79.8} & \textbf{68.8} & \textbf{81.5} \\
\bottomrule[1pt]
\end{tabular}

\vspace{2pt}
{\footnotesize\textbf{Axes dimensionalities}}
\end{minipage}
\hfill
\begin{minipage}[t]{0.48\linewidth}
\centering
\includegraphics[width=0.95\linewidth]{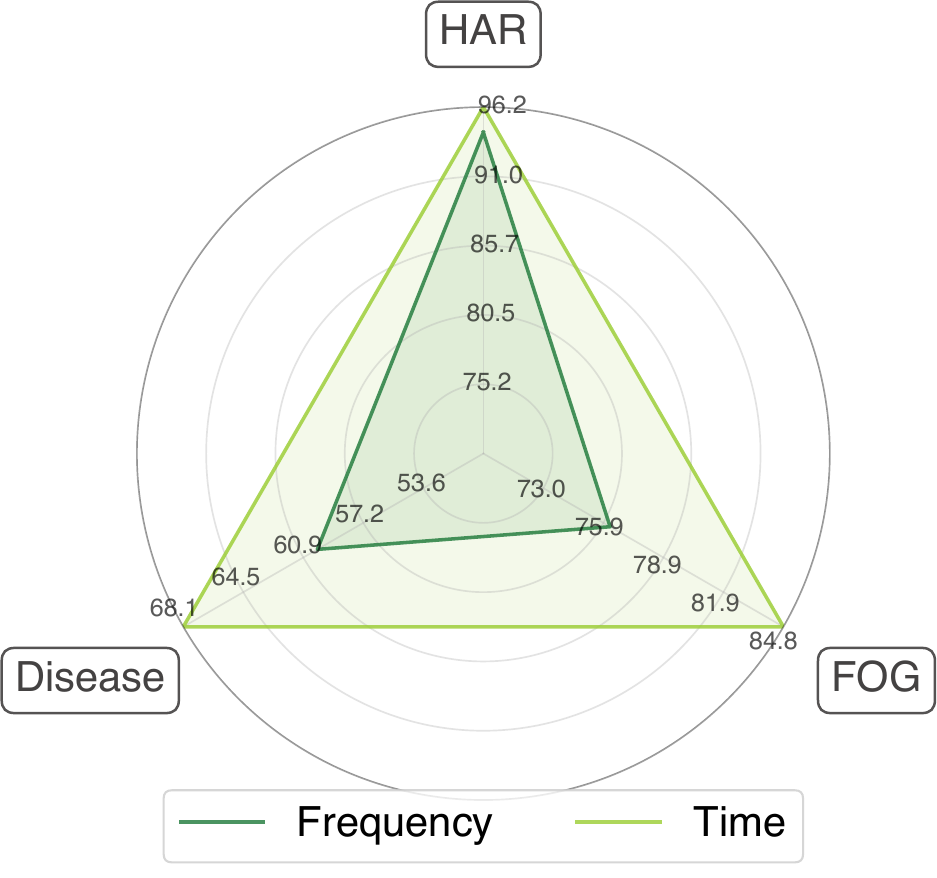}

\vspace{0pt}
\scriptsize
\setlength{\tabcolsep}{2.5pt}
\begin{tabular}{p{1cm}cccc}
\toprule[1pt]
Input & HAR & FoG & DP & Overall \\
\midrule
Freq. & 94.3 & 76.3 & 60.0 & 76.9 \\
\grayrow
Time & \textbf{96.2} & \textbf{84.8} & \textbf{68.1} & \textbf{83.0} \\
\bottomrule[1pt]
\end{tabular}

\vspace{2pt}
{\footnotesize\textbf{Representation domain}}
\end{minipage}

\vspace{-2pt}
\caption{\small{
\textbf{Control studies: axis dimensionality and representation domain.} \textbf{(Left)} Full-finetuning AUROC for uniaxial versus triaxial inputs. \textbf{(Right)} Full-finetuning AUROC for frequency-domain versus time-domain representations. Triaxial inputs preserve stronger transferable motion structure, while time-domain modeling performs better for \fog and disease prediction.
}}
\label{fig:input-ablation}
\vspace{-10pt}
\end{wrapfigure}

should be matched to task granularity: short-window activity labels favor localized motion structure, whereas gait and health tasks may benefit from longer behavioral context.

\textbf{Triaxial inputs preserve transferable motion structure.}
Following common cohort preprocessing practice, we compare full triaxial acceleration with vector magnitude, which collapses three axes into one channel. As shown in Fig. \ref{fig:input-ablation}, triaxial inputs consistently outperform uniaxial inputs across task families. This indicates that directional and orientation-dependent motion patterns carry transferable information that cannot be fully recovered after magnitude aggregation. Interestingly, \artransformer is relatively more resilient to this loss than \patchtst (Fig. \ref{fig:patch-ax}), suggesting that architecture can influence robustness to compressed input representations, but preserving raw axes remains the stronger default choice.

\textbf{Time-domain modeling better preserves gait and disease signals.}
We compare frequency-domain reconstruction with direct temporal modeling by evaluating \selfpab and \patchtst under matched settings. As shown in Fig. \ref{fig:input-ablation}, the representation domain has limited impact on \har, but time-domain modeling performs better for \fog and disease prediction. This suggests that gait events and disease-relevant movement signatures depend on precise temporal structure that may be weakened by spectrogram-style transformations. Overall, these results show that sensing choices are not secondary implementation details; they directly shape what information motion foundation models can transfer.

\vspace{-5pt}
\subsection{Multi-Stream Sensing Reveals Complementary Motion Structure}
\vspace{-3pt}

\textbf{Multiple synchronized streams add complementary motion views.}
Wearable datasets often contain richer signals than a single wrist-worn accelerometer stream, including multiple sensor modalities and body placements. We therefore ask whether representations improve when additional streams are fused, and whether a model pretrained only on wrist accelerometry can transfer to these different physical views of motion.

To separate the effects of pretraining and stream fusion, we instantiate four settings: \ding{182} supervised single-stream, \ding{183} pretrained single-stream, \ding{184} supervised multi-stream fusion, and \ding{185} pretrained multi-stream fusion. For the single-stream setting, we use the wrist or accelerometer stream when available, falling back to the most common placement or modality otherwise. For multi-stream fusion, we temporally align windows across all available streams, process each stream with a separate encoder, and concatenate the resulting embeddings for downstream classification. Supervised models are trained from scratch, while pretrained models are initialized from the same pretrained wrist-accelerometer checkpoint.

\begin{wrapfigure}{r}{0.44\textwidth}
\vspace{-10pt}
\centering
\includegraphics[width=1.0\linewidth]{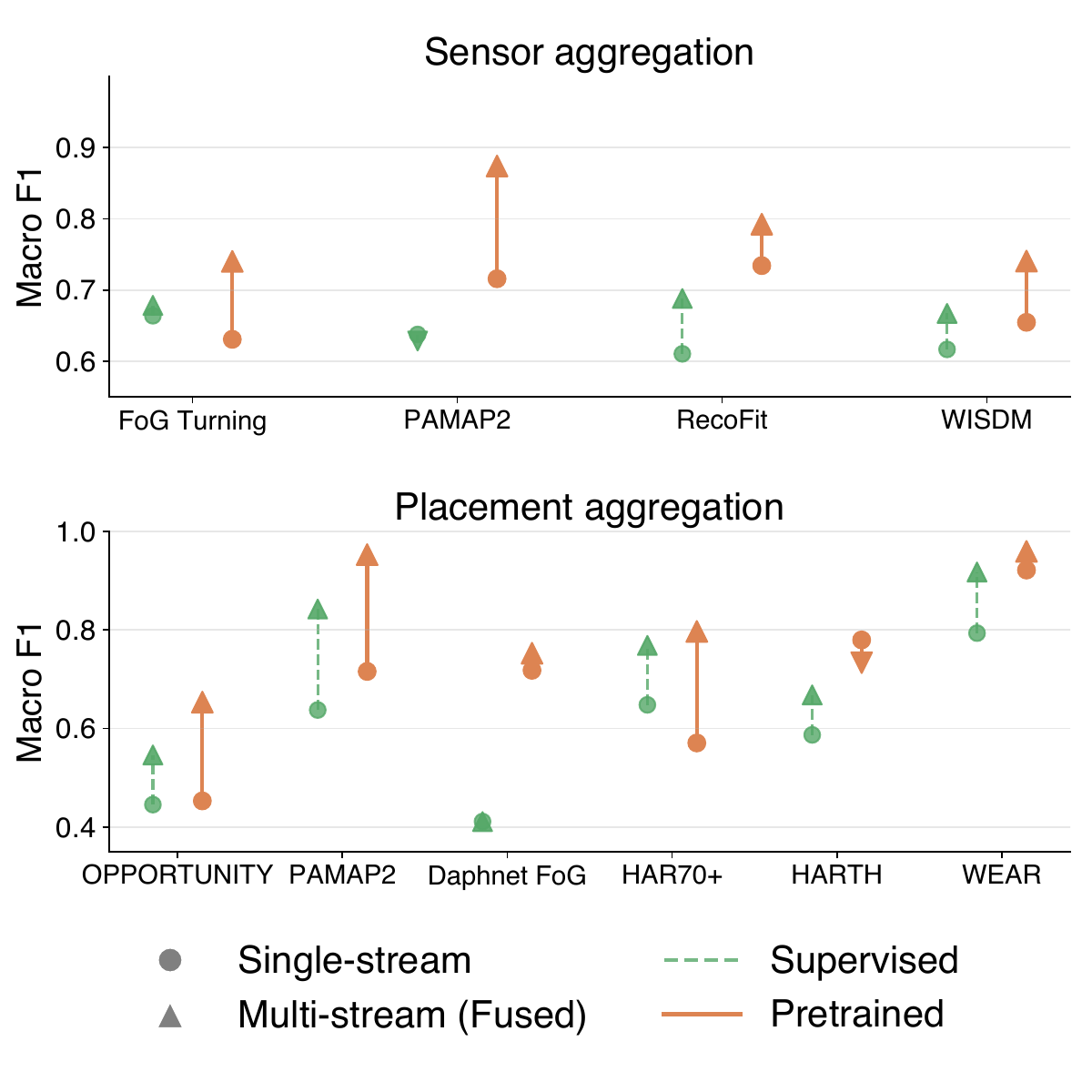}
\vspace{-20pt}
\caption{\small{\textbf{Wrist-accelerometer pretraining transfers across sensors and placements.} We report macro-F1 for default single-stream inputs and fused multi-stream inputs.}}
\vspace{-10pt}
\label{fig:sensor_placement_improvement}
\end{wrapfigure}

As shown in \figref{fig:sensor_placement_improvement}, adding streams consistently improves performance for both sensor-modality fusion and body-placement fusion. This shows that different sensors and placements provide complementary task-relevant information rather than redundant copies of the same signal. More importantly, the wrist-accelerometer pretrained model improves performance across both single-stream and multi-stream settings, even when evaluated on unseen placements or sensor modalities. This suggests that large-scale wrist pretraining learns a general representation of human motion, rather than a representation tied only to one device location or hardware channel.

\begin{figure}[t]
    \centering
    \includegraphics[width=\textwidth]{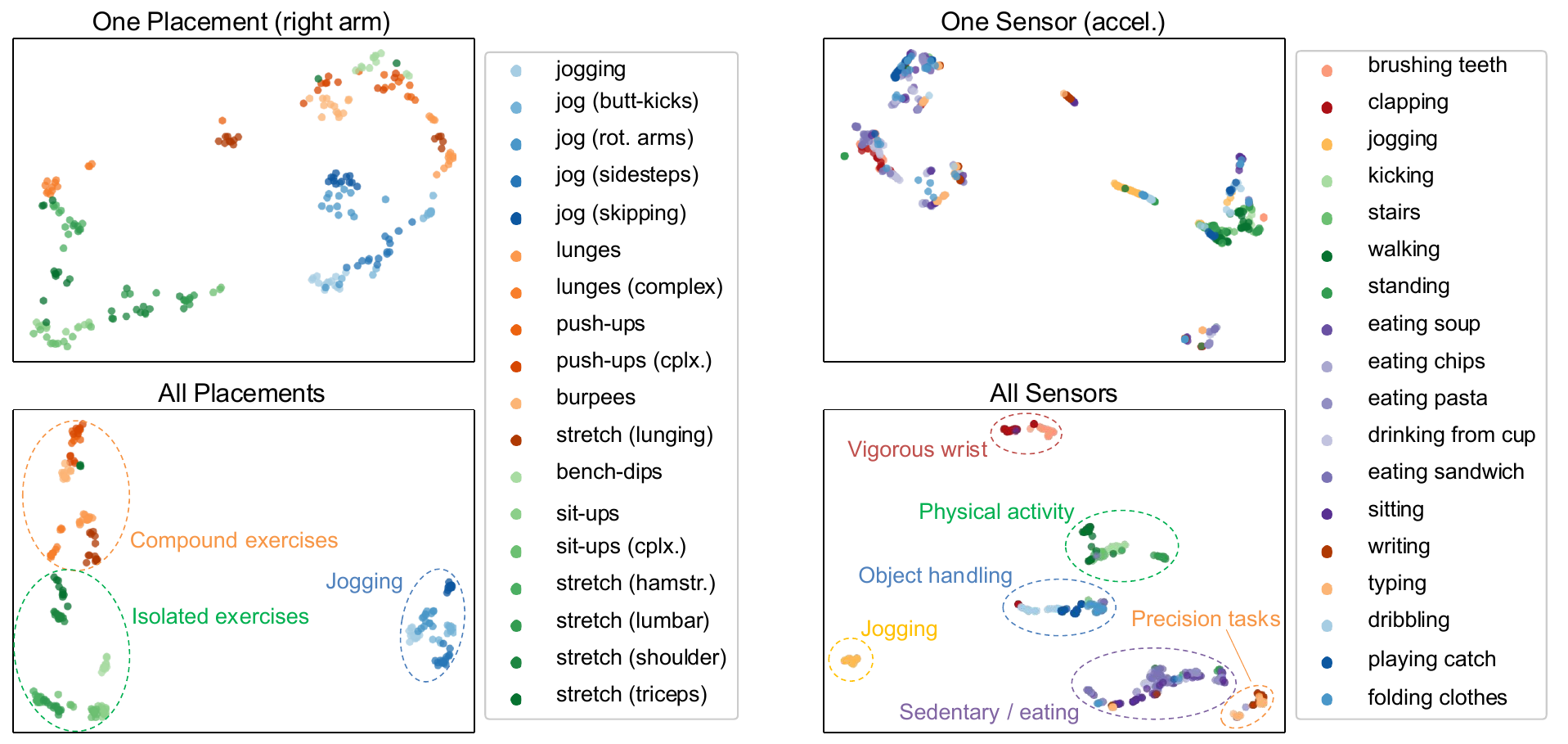}
    \caption{\small{\textbf{Multi-stream fusion improves representation geometry.} \textbf{Left:} \wear embeddings from a single right-arm placement \emph{vs.} fused multi-placement. \textbf{Right:} \wisdm embeddings from a single accelerometer stream \emph{vs.} fused multi-sensor. In both datasets, aggregating multiple streams produces more separated activity clusters.}}
    \label{fig:sensor_placement_umap}
\vspace{-8pt}
\end{figure}

\textbf{Fusion improves representation geometry.}
We further examine whether multi-stream fusion changes the structure of the learned representation space. As shown in Fig. \ref{fig:sensor_placement_umap}, embeddings from fused streams form better-separated activity clusters than embeddings from the default single stream. This pattern is visible for both \textit{placement fusion} in \wear and \textit{sensor fusion} in \wisdm, where multi-stream embeddings separate compound exercises, isolated exercises, jogging, object handling, physical activity, and sedentary behaviors more clearly. Quantitative cluster statistics in \tabref{tab:appendix_cluster_stats} further support this improved representation geometry. Together, these results suggest that multi-stream sensing provides complementary views of movement, and that pretraining helps align these views into a more separable and transferable motion representation.


%% file: tables/main-model-comp.tex
\begin{table}[t]
\centering
\small
\setlength{\tabcolsep}{6pt}
\caption{\small{\textbf{Comparisons of SSL models and supervised baselines across all \har datasets.} We report the linear probing AUROC of each model. AUPRC and full finetuning results are in Table \ref{tab:har_auprc} \& \ref{tab:appendix:har_ff}.
}}
\label{tab:har_auroc}
\vspace{3pt}
\begin{adjustbox}{width=1\textwidth,center}
\begin{tabular}{lcccccccccc}
\toprule[1.5pt]
Model & \hhar & \mhealth & \opportunity & \pamap & \recofit & \capture & \harplus & \harth & \wear & \wisdm \\
\midrule[1.5pt]
\rowcolor{gray!15}
\multicolumn{11}{l}{\emph{Supervised Models}}\\
\addlinespace[1pt]
ViT & 83.9 & 93.8 & 72.9 & 92.0 & 97.7 & 92.9 & 96.6 & 98.7 & 97.7 & 94.1 \\
CNN & 89.1 & 90.7 & 69.5 & 95.5 & 97.0 & 90.9 & 94.5 & 98.7 & 98.0 & 94.8 \\
\midrule
\addlinespace[1pt]
\midrule
\rowcolor{gray!15}
\multicolumn{11}{l}{\emph{General Pretraining Methods}}\\
\addlinespace[1pt]
\patchtstCite & \underline{93.5} & \textbf{98.2} & 89.8 & \textbf{97.4} & 98.5 & 91.6 & 95.8 & \underline{98.9} & 99.2 & \underline{95.5} \\
\artransformerCite & 88.6 & 96.0 & \textbf{92.1} & \underline{95.6} & \textbf{98.9} & \textbf{92.3} & 95.5 & \textbf{99.0} & \textbf{99.3} & 94.8 \\
\argruCite & 89.2 & 96.0 & 89.4 & 95.0 & \underline{98.7} & 91.3 & 95.3 & 98.5 & \underline{99.2} & 94.7 \\
\dinoCite & 73.4 & 87.2 & 87.8 & 82.7 & 98.1 & 92.0 & 95.4 & 98.9 & 98.5 & 85.2 \\
\simclrCite & 68.6 & 90.0 & \underline{90.0} & 86.8 & 97.8 & \underline{92.2} & \underline{95.9} & 98.6 & 98.0 & 86.4 \\
\midrule
\rowcolor{gray!15}
\multicolumn{11}{l}{\emph{Domain-specific Methods}}\\
\addlinespace[1pt]
\sslwearablesCite & 90.0 & 89.9 & 73.1 & 91.8 & 91.0 & 87.0 & 85.4 & 92.0 & 88.5 & 90.2 \\
\relconCite & 87.8 & \underline{96.3} & 85.3 & 94.5 & 97.0 & 89.9 & \textbf{95.9} & 98.0 & 97.9 & 90.5 \\
\selfpabCite & \textbf{96.8} & 91.8 & 74.7 & 95.4 & 97.4 & 89.7 & 93.4 & 98.6 & 97.7 & \textbf{96.7} \\
\bottomrule[1.5pt]
\end{tabular}
\end{adjustbox}
\vspace{-3pt}
\end{table}

%% file: tables/main-model-fog.tex
\begin{figure}[t] 
\centering
\begin{minipage}[t]{0.49\textwidth}
\vspace{0pt}
\centering
\small
\parbox[t][4.8em][t]{\linewidth}{
\captionof{table}{\small{\textbf{Comparison of SSL models and supervised baselines across all \fog tasks.} We report the linear probing AUROC of each model. AUPRC and full finetuning results are in Table \ref{tab:fog_auprc} \& \ref{tab:appendix:fog_ff}.
}}
\label{tab:fog_auroc}}
\setlength{\tabcolsep}{3pt}
\begin{adjustbox}{width=\linewidth}
\begin{tabular}{lccc}
\toprule[1.2pt]
Model & \fogturning & \odayfog & \daphnet \\
\midrule[1.2pt]

\rowcolor{gray!15}
\multicolumn{4}{l}{\emph{Supervised Models}}\\
\addlinespace[1pt]

ViT & 63.6 & 68.0 & 83.5 \\
CNN & 51.2 & 80.5 & 87.6 \\

\midrule
\addlinespace[1pt]
\midrule

\rowcolor{gray!15}
\multicolumn{4}{l}{\emph{General Pretraining Methods}}\\
\addlinespace[1pt]

\patchtstCite & 85.7 & 65.4 & \textbf{95.0} \\
\artransformerCite & 80.2 & 58.6 & 94.3 \\
\argruCite & 83.0 & 71.6 & 93.4 \\
\dinoCite & 61.4 & \textbf{83.2} & 89.2 \\
\simclrCite & \underline{91.4} & 66.7 & \underline{94.8} \\

\midrule

\rowcolor{gray!15}
\multicolumn{4}{l}{\emph{Domain-specific Methods}}\\
\addlinespace[1pt]

\sslwearablesCite & \textbf{91.6} & 62.0 & 71.4 \\
\relconCite & 87.5 & \underline{78.2} & 93.0 \\
\selfpabCite & 80.4 & 55.0 & 92.5 \\

\bottomrule[1.2pt]
\end{tabular}
\end{adjustbox}
\end{minipage}
\hfill
\begin{minipage}[t]{0.48\textwidth}
\vspace{0pt}
\centering
\small
\parbox[t][4.8em][t]{\linewidth}{
\captionof{table}{\small{\textbf{Comparison of SSL models and supervised baselines across disease prediction tasks.} We report the AUROC of MIL on frozen backbone results for each model. Full results are in Table \ref{tab:disease_auprc}.
}}
\label{tab:disease_auroc}}
\setlength{\tabcolsep}{2.3pt}
\begin{adjustbox}{width=\linewidth}
\begin{tabular}{lccccc}
\toprule[1.2pt]
Model & Depr. Sev. & Depr. & Diabetes & PD & Sleep \\
\midrule[1.2pt]

\rowcolor{gray!15}
\multicolumn{6}{l}{\emph{Supervised Models}}\\
\addlinespace[1pt]

ViT & 49.5 & 57.9 & 49.4 & 45.4 & 38.3 \\
CNN & 51.7 & 59.5 & 61.9 & 53.7 & 54.0 \\

\midrule
\addlinespace[1pt]
\midrule

\rowcolor{gray!15}
\multicolumn{6}{l}{\emph{General Pretraining Methods}}\\
\addlinespace[1pt]

\patchtstCite & 55.0 & \underline{77.0} & 53.8 & 72.8 & 59.6 \\
\artransformerCite & 58.4 & \textbf{81.3} & 67.7 & \textbf{75.0} & 61.7 \\
\argruCite & 60.6 & 76.2 & 53.2 & 72.4 & 61.1 \\
\dinoCite & \underline{62.9} & 75.8 & 74.4 & 54.5 & \underline{76.5} \\
\simclrCite & 58.0 & 70.7 & \textbf{78.8} & 51.3 & \textbf{77.2} \\

\midrule

\rowcolor{gray!15}
\multicolumn{6}{l}{\emph{Domain-specific Methods}}\\
\addlinespace[1pt]

\sslwearablesCite & \textbf{65.8} & 60.5 & \underline{78.2} & \underline{74.4} & 65.3 \\
\relconCite & 57.8 & 72.6 & 70.5 & 51.0 & 56.8 \\
\selfpabCite & 52.0 & 65.3 & 59.6 & 60.3 & 58.5 \\

\bottomrule[1.2pt]
\end{tabular}
\end{adjustbox}
\end{minipage}
\vspace{-5pt}
\end{figure}

%% file: sections/5_discussion.tex
\vspace{-4pt}
\section{Discussion}
\vspace{-4pt}

\textbf{Limitations.}
While \ours provides a controlled exploration of wearable motion FMs, several limitations remain. First, not all pretraining objectives may respond similarly to each design axis. Our detailed ablations focus on top-performing representative methods, but contrastive, self-distillation, reconstructive, and predictive objectives may each have different sensitivities to sampling rate, window length, placement, modality, and scale. A finer objective-specific sweep may reveal additional design patterns. Second, our downstream evaluations are classification-based, whereas real-world wearable applications also include continuous health prediction, biomarker regression, longitudinal trajectory modeling, and event forecasting. Extending \ours to these settings is an important next step for understanding how motion FMs support broader health and behavior modeling.

\textbf{Conclusion.}
We present \ours, a fully open exploration of wearable motion FMs that addresses fragmentation across the full lifecycle of motion representation learning.
Across more than 1,000 models and evaluations on 18.2M hours of accelerometry data from 115,000+ individuals, we find that self-supervised pretraining consistently improves transfer over supervised training from scratch, but no single objective dominates across all task families. Our analyses further show that data representation, temporal resolution, placement, and pretraining scale can be as important as model architecture itself. Collectively, \ours provides state-of-the-art recipes for diverse downstream tasks and serves as a comprehensive, practical, and open cookbook for motion representation learning.

%% file: sections/appendix.tex
\section{Experimental Scope, Data Regimes, and Evaluation Setup}
\label{app:data_eval}

\subsection{Experimental axes}
To make the design space explicit, \ours evaluates wearable motion foundation models across the major choices that can be made in wearables pipelines: self-supervised objective, model capacity, sampling frequency, window length, sensor axis dimensionality, sensor modality, body placement, and input representation domain. \tabref{tab:inertia1_breadth} summarizes the resulting grid of experimental axes used throughout the paper. The list of Device Placements in order of badges are: wrist, hand, arm, back, chest, hip, knee, leg, shank, ankle.
\input{tables/main_breadth}

\subsection{Pretraining data}
Our pretraining corpus is mainly derived from the National Health and Nutrition Examination Survey (\nhanes) accelerometry cohort from year 2011-2014. Participants wore a physical activity monitor continuously for seven days. Raw triaxial accelerometer signals were collected at 80 Hz, low-pass filtered using a 4th-order Butterworth filter with a 10 Hz cutoff, and downsampled to 20 Hz. The processed data are stored in ten-minute parquet segments, comprising 14,688 subjects and approximately 2.47 million subject-hours of recordings. This dataset provides large-scale, high-resolution unlabeled motion data suitable for representation learning. 

In addition, we utilize the \ukb accelerometer dataset for data scaling experiments, including cross-dataset mixing with \nhanes. Approximately 100,000 participants wore a physical activity monitor continuously for seven days. Raw 100 Hz triaxial accelerometer signals were auto-calibrated to local gravity, gravity-corrected, and low-pass filtered using a 4th-order Butterworth filter with a 20 Hz cutoff. The processed Euclidean norm data available to us are stored as pre-computed five-second epoch time-series files, with non-wear periods imputed using time-of-day averages. Filtered for subjects meeting a minimum 72-hour wear-time threshold, this cohort comprises approximately 15.7 million subject-hours of recordings, further expanding the scale and diversity of our unlabeled pretraining corpus.

\subsection{Downstream datasets and task families}
We evaluate learned representations on thirteen public \har and \fog benchmarks, spanning diverse sensor placements, sampling frequencies, and activity label sets. In addition, we evaluate on five patient-level \dd tasks from \nhanes and two patient-level \dd tasks derived from \ukb ICD10 codes, using multiple-instance pooling over windows from each subject. Train/val/test splits are selected on the subject level for all datasets. A comprehensive summary of downstream datasets is provided in \tabref{tab:appendix:data}.
\input{tables/appendix_downstream_data}

\subsection{Training and evaluation protocol}
\label{app:train_eval_protocol}
Unless otherwise specified, models are pretrained for $170k$ optimization steps using AdamW on 4 H200 GPUs. Default controlled experiments use $30s$ windows of 20 Hz triaxial accelerometer data from wrist-worn devices, or the closest available placement when a dataset does not include a wrist-worn IMU. The default encoder capacity is the Small setting, corresponding to approximately 5--8M parameters, with Medium ($\sim$30M) and Large ($\sim$100M) variants used for scaling analyses. Downstream evaluation uses linear probing, where the pretrained encoder is frozen and only a lightweight classifier is trained, and full fine-tuning, where all model parameters are updated end-to-end. 

For patient-level \dd in \nhanes, we use multiple-instance learning by sampling 1024 windows from each subject at predetermined times of the day and pooling their embeddings into a subject-level prediction. Labels are derived from \nhanes survey reports and medications data. Parkinson's disease, depression (binary), and diabetes are inferred through use of anti-Parkinson's agents, antidepressants, and insulin, respectively. Additionally, we report the average metric across 5 run seeds for \dd to reduce performance variance due to low positive sample size. See Appendix \ref{app:ukb_disease_prediction} for details on \ukb \dd.

\section{Pretraining Algorithms and Model Implementations}
\label{app:pretraining_algorithms}

Unless otherwise specified, method implementations use the shared default preprocessing and training configuration: 30-second windows sampled at 20 Hz, triaxial accelerometer input, a batch size of 1024, AdamW optimization with learning rate $2\times10^{-4}$ and weight decay 0.04, and 5 pretraining epochs. All pretraining runs utilized a learning rate scheduler with linear warmup and cosine decay. Downstream evaluation uses the linear-probing and full-finetuning protocols described in Appendix~\ref{app:train_eval_protocol}. The probing and finetuning runs use learning rate $1\times10^{-3}$ and weight decay 0.01. Size-controlled experiments use the corresponding small, medium, and large configuration presets when applicable.

\subsection{General self-supervised methods}

\paragraph{\patchtst.}
We implement \patchtst as a masked patch-reconstruction model for accelerometer windows~\cite{nie2023patchtst}. The input window is divided into temporal patches and a random subset of patches is masked before being passed to the \patchtst encoder. We use patch length 10, stride 10, and mask ratio 0.4. The model predicts the original unmasked patch values, and the pretraining loss is mean-squared error evaluated only on masked patches. For downstream evaluation, we use the mean-pooled \patchtst backbone representation. In model-size experiments, the preset-controlled \patchtst sizes are: small $(d=512, L=4, H=4, d_{ff}=1024)$, medium $(d=768, L=6, H=8, d_{ff}=2048)$, and large $(d=1024, L=8, H=16, d_{ff}=4096)$.

\paragraph{\artransformer.}
The \artransformer is trained with a causal next-patch prediction objective. The input is divided into non-overlapping temporal patches of size 10 by default. The pretraining loss is mean-squared error between the predicted and target next patches. The model uses a linear patch projection, a causal Transformer encoder, and a linear prediction head. For downstream evaluation, the default representation is generated by mean pooling over all final layer positions. In model-size experiments, the preset-controlled \artransformer sizes are: small $(d=512, L=4, H=4, d_{ff}=1024)$, medium $(d=768, L=6, H=8, d_{ff}=2048)$, and large $(d=1024, L=8, H=16, d_{ff}=4096)$.

\paragraph{\argru.}
The \argru uses the same next-patch prediction objective as \artransformer, but replaces the Transformer encoder with a recurrent GRU. The default patch length is 10 samples. Each patch is projected into the GRU hidden space, the GRU processes the patch sequence causally, and a linear head predicts the next patch. The pretraining loss is mean-squared error on next-patch prediction. For downstream evaluation, the default representation is the mean of the GRU hidden states over time.

\paragraph{\dino.}
We adapt \dino to wearable motion using a student--teacher self-distillation setup~\cite{caron2021emerging}. Both student and teacher use the same ViT-style motion encoder, and the teacher parameters are initialized from the student and then updated as an exponential moving average of the student parameters. Under the default configuration, both encoders use a patch size of 10 on the temporal axis for input embedding. The base augmentations are adapted from genertic augmentation strategies such as adding noises and masking inputs. Specifically, we use jitter with standard deviation 0.02, scaling with standard deviation 0.1, time masking with ratio sampled from $[0.3,0.6]$, and channel dropout over 0 or 1 channels. The \dino loss matches all student views to the teacher outputs from the global views. For downstream evaluation, the implementation uses the raw backbone representation from the student encoder by default and discards the \dino head.

\paragraph{\simclr.}
We adapt \simclr to wearable motion by generating two augmented views of each input window and training a shared encoder with a contrastive objective~\cite{chen2020simple}. The default encoder is the same ViT-style motion encoder used in \dino, with patch size of 10. Training uses the NT-Xent loss with temperature 0.1. We use the same genertic augmentation strategies as \dino by default. For downstream evaluation, the implementation returns the raw encoder representation and discards the \simclr projection head.
\subsection{Domain-specific self-supervised methods}

\paragraph{\sslwearables.}
Following the original implementation by \cite{yuan2024self}, we frame the \sslwearables pretraining objective as a three-task temporal transformation prediction problem. For each input window, the model constructs views corresponding to arrow-of-time, permutation, and time-warping tasks. We also follow the official codebase to dictate the transformation parameters and loss calculations. For downstream evaluation, we follow standard procedure by extracting features directly from the raw ResNet encoder and discarding the self-supervised prediction heads.

\paragraph{\relcon.}
Following the original implementation by \cite{xu2025relcon}, we employ a two-stage self-supervised approach that replaces fixed distance metrics with a learned reconstructability criterion. In the first stage, a distance model is trained to identify motion-specific motifs across accelerometer windows. In the second stage, this model is frozen and used to rank candidate windows for a relative contrastive loss. We strictly adhere to the official codebase to implement the loss formulation and default hyperparameters over a ResNet1D encoder.

\paragraph{\selfpab.}
We implement \selfpab as masked reconstruction in a spectrogram representation~\cite{logacjov2024selfpab}. The module computes the short-time Fourier transform internally. The default STFT uses $n_{\mathrm{fft}}=128$, hop length 64, no centering, and magnitude-only input without phase to adjust for our longer window length. The masking procedure alters both time and frequency regions, and the reconstruction objective is masked L1 loss by default. For downstream evaluation, the model converts the raw window to its STFT representation without masking, applies through the model encoder, and returns the mean-pooled frame representation.

\subsection{Supervised baselines}

\paragraph{CNN.}
The CNN baseline implementation uses a ResNet-style 1D convolutional encoder with input channels equal to the selected sensor axes. The method is trained by the downstream evaluation code with the task-specific supervised objective.
\paragraph{ViT.}
The ViT baseline divides each motion window into temporal patches and processes them with a 1D ViT encoder. The patch size is default to 10. As with the CNN baseline, the method is trained by the downstream evaluation code with the task-specific supervised objective, rather than by a self-supervised pretraining loss.

\subsection{Model capacity definitions}
We report model capacity using the Small, Medium, and Large settings used in the scaling experiments. \tabref{tab:appendix_model_sizes} lists the parameter counts for each backbone.
\input{tables/appendix_model_sizes}

\section{Supplementary Metrics for Main Results}
\label{app:supp_main_results}

The main paper reports linear probing AUROC as the primary metric because it allows comparison across all task families. Since several downstream datasets are class-imbalanced, we also report both linear probing and full finetuning AUPRC and task-specific breakdowns here (\tabref{tab:har_auprc}, \tabref{tab:fog_auprc}, \tabref{tab:disease_auprc}). In additon, we also include full finetuning AUROC across \har (\tabref{tab:appendix:har_ff}) and \fog (\tabref{tab:appendix:fog_ff}) tasks. Finally, we include the raw AUROC and AUPRC across all axes ablations discussed in the main paper (\tabref{tab:ablation_auroc_all}, \tabref{tab:ablation_auprc_all}).

\input{tables/appendix-modelcomp-auprc}
\input{tables/appendix-model-dp}

\input{tables/appendix-model-har-ff}

\section{Supplementary Scaling Analyses}
\label{app:supp_scaling}

\subsection{Parameter scaling under fixed pretraining data}
\begin{figure}[t]
    \centering
    \includegraphics[width=\textwidth]{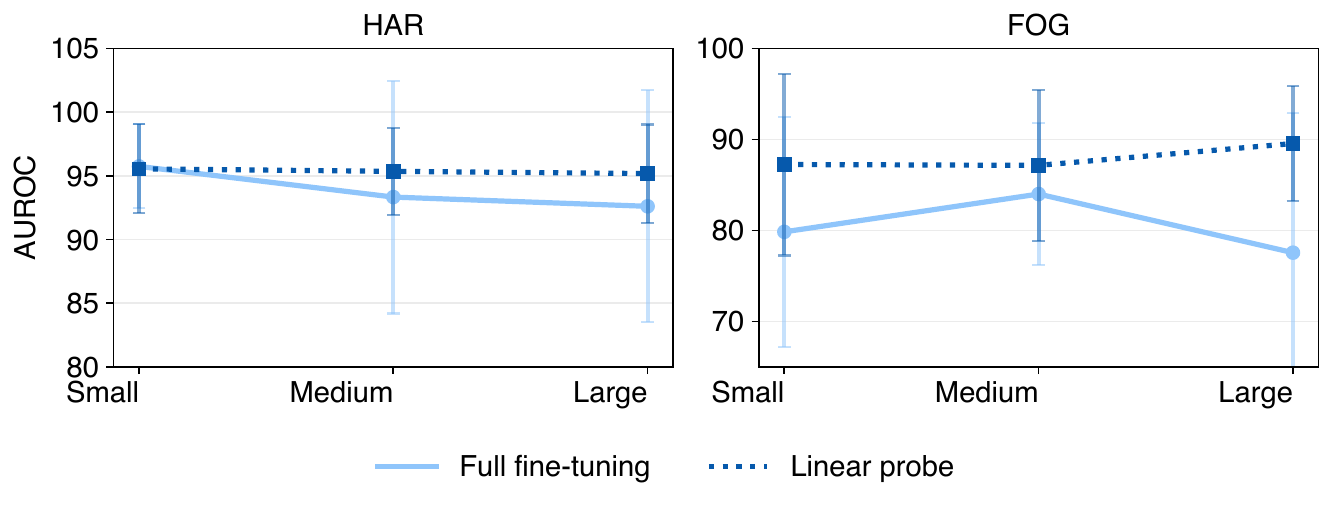}
    \caption{\textbf{Parameter scaling under fixed pretraining data.}
    Controlled ablations using \artransformer. We vary encoder size and compare
    full fine-tuning against linear probing while holding the pretraining corpus
    and input configuration fixed.}
    \label{fig:mscaling}
\end{figure}

We investigate scaling behavior by varying model capacity (Small $<$ 10M, Medium $\sim$30M, Large $\sim$100M parameters) and pretraining corpus size. \figref{fig:mscaling} compares full fine-tuning against linear probing for \artransformer under the same fixed pretraining data and input configuration. The full-fine-tuning plots show a slight degradation in performance, suggesting that larger encoders alone are not a reliable recipe without corresponding gains in data scale, diversity, or task-specific optimization. All error bars in scaling plots reflect the standard deviation across specific disease tasks and trials, calculated via function call. Standard deviations were calculated using the Python function \verb|statistics.stdev()|.

\input{tables/appendix_ablations}
\subsection{PatchTST capacity under linear probing and full fine-tuning}
\begin{figure}[t]
    \centering
    \includegraphics[width=\textwidth]{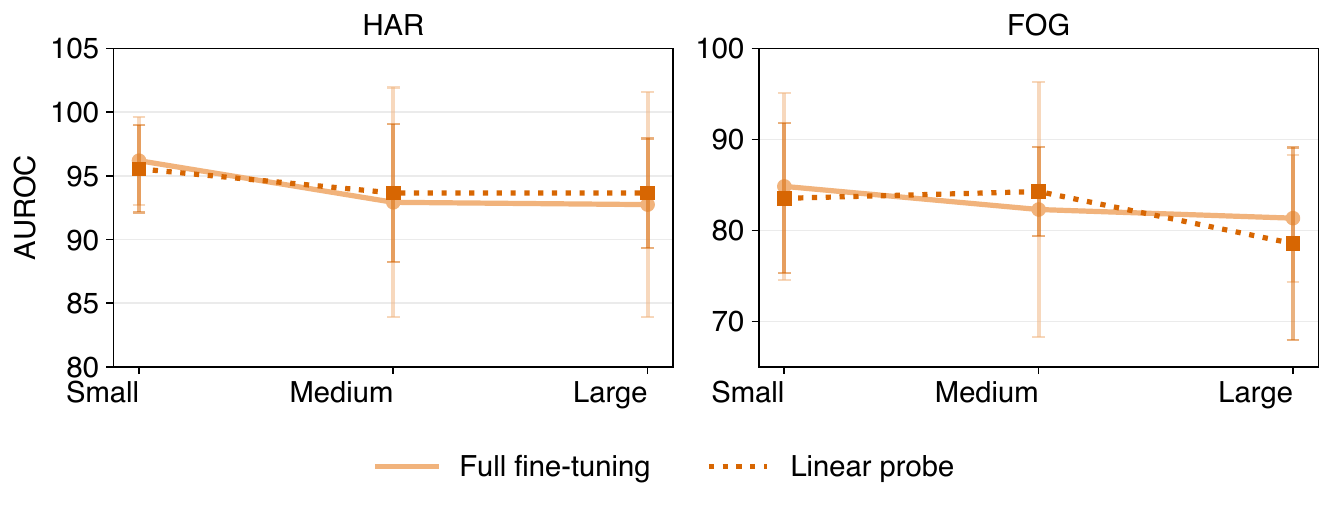}

    \caption{\textbf{Additional Model Capacity Ablations.} Controlled ablations using \patchtst; we vary the capacity of the encoder and style of finetuning while holding the rest of the protocol fixed.}
    \label{fig:patchmscaling}
\end{figure}

\figref{fig:patchmscaling} provides the capacity ablation for \patchtst. This analysis is included as a robustness check because the main text focuses primarily on \artransformer for controlled ablations. Across both pretrained  models evaluated, we find that scaling model parameters alone does not yield meaningful improvements in representation. On the contrary, in the case of finetuning the entire model backbone, both \artransformer and \patchtst performances begin to degrade as the models overfit limited downstream training data. We find, however, that \artransformer, performance remains comparable as model capacity increases, regardless of finetuning method, suggesting that \artransformer scales more gracefully than does \patchtst.

\input{tables/appendix_ablation_auroc}
\input{tables/appendix_ablation_auprc}

\subsection{Disease prediction scaling protocol}
\label{app:ukb_disease_prediction}
The \ukb disease scaling experiments use a scaling-specific pretraining and evaluation protocol. In this setting, we separate two sources of additional unlabeled data: the number of pretraining individuals and the number of motion segments per individual. For these experiments, we train an \artransformer using 2-hour windows of accelerometer  data downsampled to 0.2\,Hz, corresponding to 1440 time steps per window. Each setting is pretrained for two epochs, and all model settings are checked to ensure normal training behavior. The pretraining split contains approximately 80\% of eligible \ukb patients, or roughly 80k patients, while the remaining eligible patients are held out from pretraining and used for downstream disease evaluation.

For each disease trait, we construct a case-control cohort from the held-out patients, using as close to a 1:5 case-control ratio as possible. This cohort is then split at the patient level into train, validation, and test folds for downstream multiple-instance learning.

For downstream evaluation, each patient-day is treated as an independent bag. A patient-day consists of 12 consecutive 2-hour windows, giving up to seven bags per patient. All bags from the same patient share the same disease label and are assigned to the same train, validation, or test fold to prevent leakage across days. The frozen encoder embeds each 2-hour window independently, and the MIL classifier is trained on bags of 12 window embeddings. Performance is measured using AUROC on the held-out test set, with 95\% bootstrap confidence intervals computed using 1000 resamples.

The three scaling sweeps vary the number of pretraining patients, the number of windows available per patient, and the composition of the pretraining corpus. The patient-scaling sweep uses 1\%, 10\%, and 100\% of the pretraining patients. The segment-scaling sweep caps the number of windows per patient at 1, 9, or 84. The mixing sweep compares \nhanes-only pretraining, matched \nhanes--\ukb mixtures, and full \ukb inclusion.

\subsection{Disease prediction scaling with additional UK Biobank targets} 

\begin{figure}[t]
    \centering
        \includegraphics[width=\textwidth]{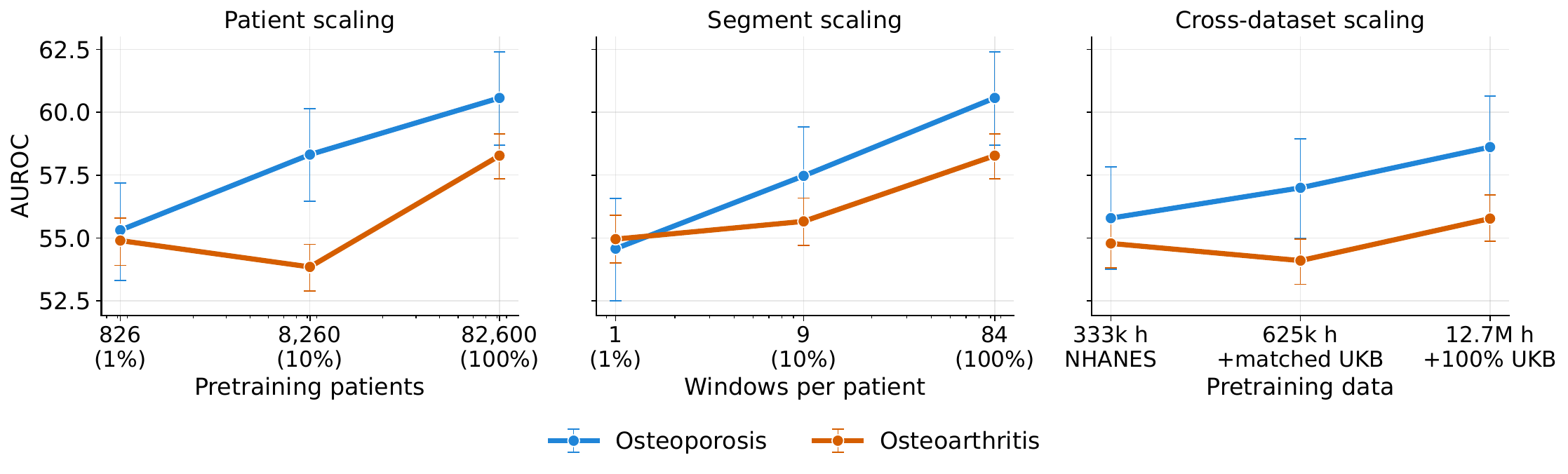}
    \caption{\textbf{\ukb disease scaling at 0.2\,Hz.} We evaluate how downstream AUROC changes as pretraining varies by number of individuals, number of segments per individual, and mixture of \nhanes and \ukb pretraining data. Individual and segment scaling both improve osteoporosis and osteoarthritis prediction, while adding in-domain \ukb data to \nhanes pretraining provides a modest additional gain.}
    \label{fig:ukb_scaling_all}
\end{figure}

\figref{fig:ukb_scaling_all} additionally visualizes performance on osteoarthritis prediction for the \ukb scaling experiments. We observe trends similar to osteoporosis, indicating that increasing the number of pretraining individuals, the number of windows per individual, and the diversity of the pretraining corpus can improve disease-relevant wearable representations across multiple targets. This further suggests that wearable disease representations benefit from both population diversity and repeated behavioral sampling within individuals.

\section{Supplementary Controlled Configuration Studies}
\label{app:controlled_config}

\subsection{Sampling frequency and window length with PatchTST}
\begin{figure}[t]
    \centering

    \begin{subfigure}[t]{0.3\textwidth}
        \centering
        \includegraphics[width=0.98\linewidth]{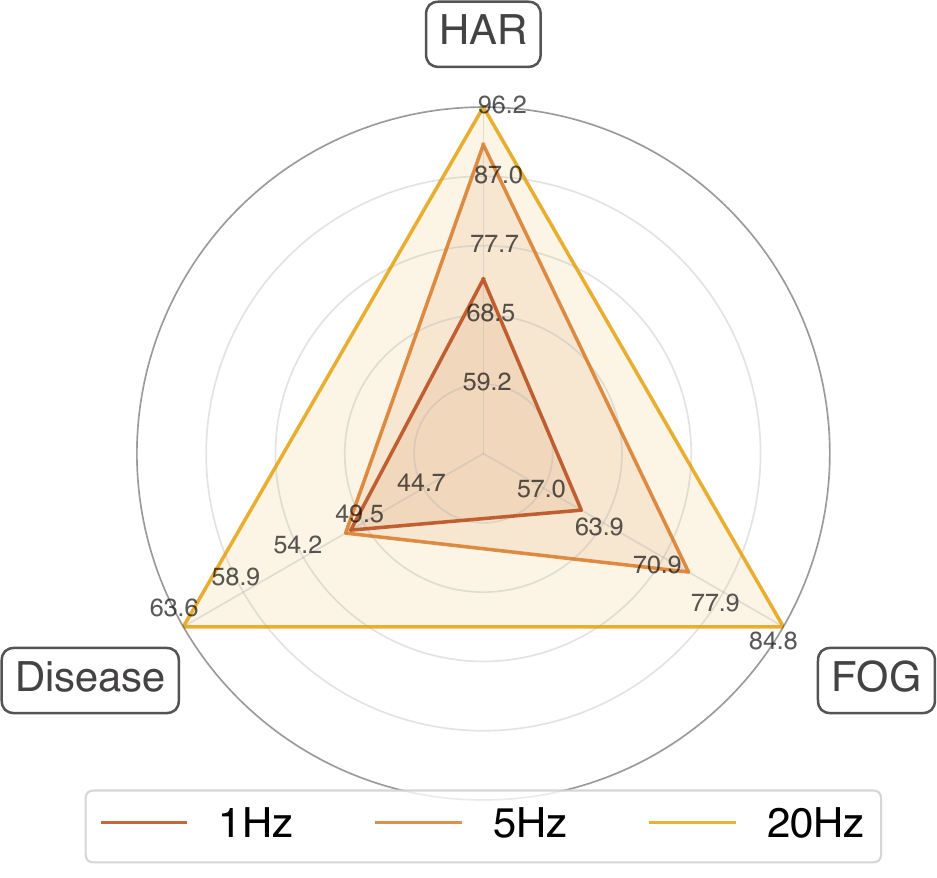}
        \vspace{0pt}

        \scriptsize
        \setlength{\tabcolsep}{2.5pt}
        \begin{tabular}{p{1cm}cccc}
        \toprule[1pt]
        Freq. & HAR & FoG & DP & Overall \\
        \midrule
        1Hz & 73.2 & 61.4 & 50.4 & 61.7 \\
        5Hz & 91.2 & 73.8 & 50.9 & 72.0 \\
        20Hz & \textbf{96.2} & \textbf{84.8} & \textbf{63.6} & \textbf{81.6} \\
        \bottomrule[1pt]
        \end{tabular}

        \caption{Sampling frequency}
        \label{fig:axes_freq}
    \end{subfigure}
    \hfill
    \begin{subfigure}[t]{0.3\textwidth}
        \centering
        \includegraphics[width=0.98\linewidth]{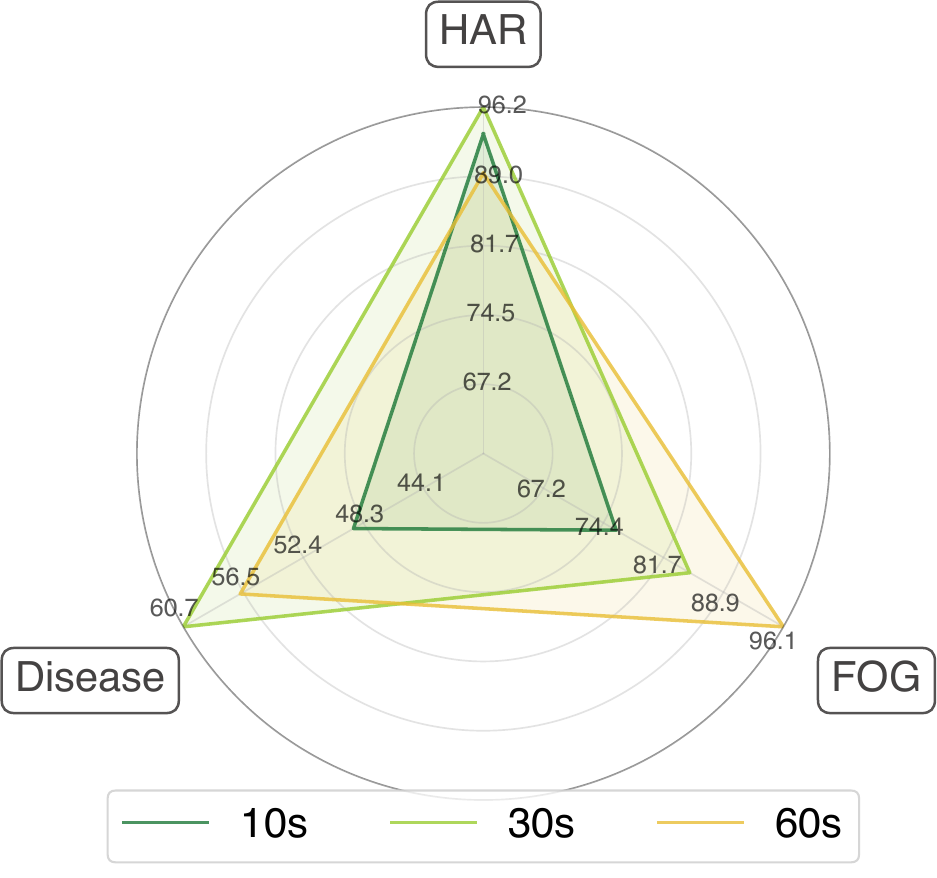}
        \vspace{0pt}

        \scriptsize
        \setlength{\tabcolsep}{2.5pt}
        \begin{tabular}{p{1cm}cccc}
        \toprule[1pt]
        Window & HAR & FoG & DP & Overall \\
        \midrule
        10s & 93.4 & 76.0 & 48.9 & 72.8 \\
        30s & \textbf{96.2} & 84.8 & \textbf{60.7} & 80.6 \\
        60s & 89.3 & \textbf{96.1} & 56.7 & \textbf{80.7} \\
        \bottomrule[1pt]
        \end{tabular}

        \caption{Window length}
        \label{fig:axes_window}
    \end{subfigure}
    \hfill
    \begin{subfigure}[t]{0.3\textwidth}
        \centering
        \renewcommand{\arraystretch}{1.35}
        \includegraphics[width=0.98\linewidth]{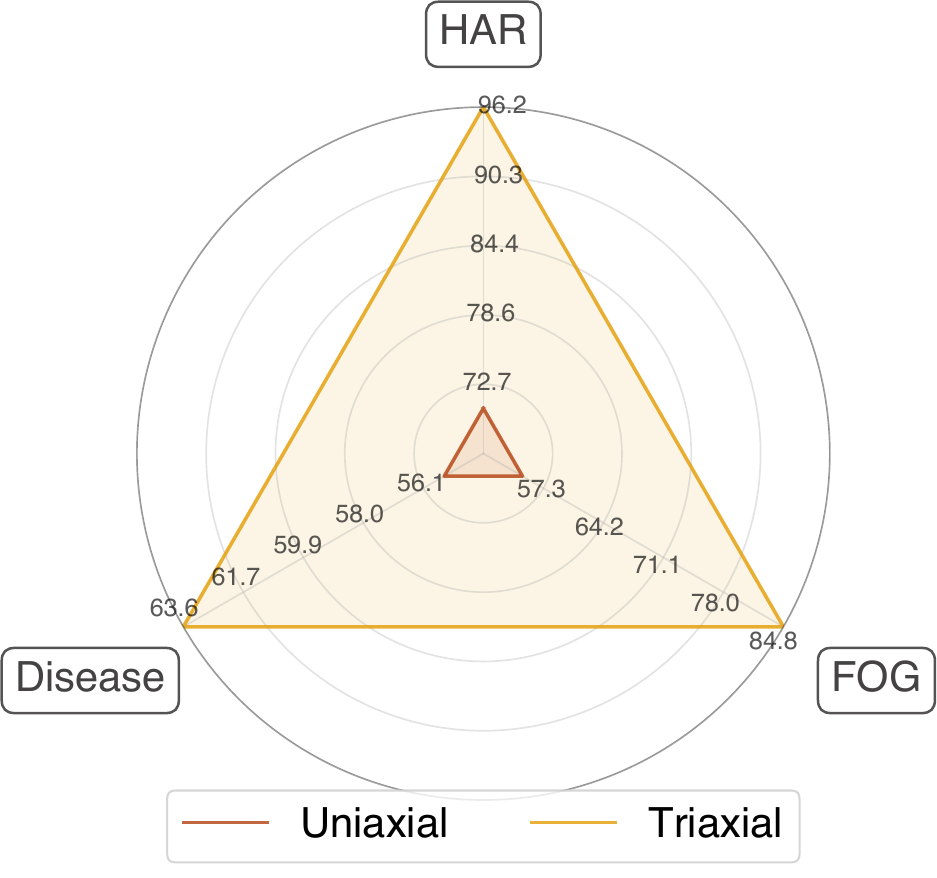}
        \scriptsize
        \setlength{\tabcolsep}{2.5pt}
        \begin{tabular}{p{1cm}cccc}
        \toprule[1pt]
        Axes & HAR & FoG & DP & Overall \\
        \midrule
        Uniaxial & 70.7 & 54.9 & 55.4 & 60.3 \\
        Triaxial & \textbf{96.2} & \textbf{84.8} & \textbf{63.6} & \textbf{81.6} \\
        \bottomrule[1pt]
        \end{tabular}
        \caption{Axis dimensionality}
        \label{fig:patch-ax}
    \end{subfigure}

    \caption{\textbf{Sensitivity to input and representation choices.}
    Controlled ablations using \patchtst. (a)(b) We vary sampling frequency and window length while holding the rest of the protocol fixed. (c) We compare performance of \patchtst trained on uniaxial inputs vs. triaxial inputs. AUROC is reported across task families.}
    \label{fig:patch_experimental_axes}
\end{figure}

The main text uses \artransformer as the primary backbone for controlled configuration studies due to its strong and balanced performance across tasks in both linear probing and full finetuning. To test whether the same qualitative trends hold for another architecture, \figref{fig:patch_experimental_axes} repeats the sampling-frequency and window-length ablations with \patchtst. Similar to the results with \artransformer in the main text, the highest temporal resolution of 20 Hz provides the greatest benefit to downstream performance, while window size effects are weaker and more task-dependent. Strikingly, model performance degrades significantly between 5Hz and 1Hz input streams. Even so, accounting for this degradation, the models pretrained at 1Hz are still able to perform sufficiently in \har classification. This finding is critical for wearable foundation models as many large cohort studies of data report only aggregate metrics or low frequency data, potentially erasing important movement information from the data. Across window size, we find less stark of a contrast; increasing context does not yield significant performance gains. We note that this saturation of performance should not be attributed to the cleanliness of the data, i.e. recordings collected of a pure action, as our window cleaniness check reports an average >80\% label purity for all of our \har and \fog splits. Detailed results are reported in Table \ref{tab:ablation_auroc_all} and Table \ref{tab:ablation_auprc_all}.

\subsection{Sensor-axis dimensionality across additional backbones}

Similar to the experiment in \figref{fig:input-ablation}, we test for qualitative trends in sensor-axis dimensionality across additional model architectures in \figref{fig:patch_experimental_axes} (\patchtst). Triaxial inputs generally preserve useful orientation and movement structure that can be lost when reducing the signal to one channel. Both models substantially improve with triaxial input data, with \patchtst appearing to collapse when given only a single axis representation of the data.



\input{tables/appendix_cluster_stats}
\section{Representation Geometry and Cluster Analysis}
\label{app:geometry}

To study placement robustness—and specifically whether wrist-based pretraining extrapolates to other sensors or placements—we finetune separate models using data from specific sensors (gyroscope, magnetometer) or body locations (waist, chest, leg, ankle) after pretraining on wrist data and train a linear classifier on concatenated embeddings for each downstream dataset with the additional available data streams. The main text interprets improvements from synchronized multi-stream fusion using representation geometry and cluster structure. \tabref{tab:appendix_cluster_stats} reports the corresponding cluster-quality metrics averaged across all datasets from \figref{fig:sensor_placement_improvement}, providing quantitative support for the qualitative embedding visualizations and cluster interpretations discussed in the main paper. The noise control row displays cluster metrics when concatenating random normal vectors of the same dimensionality as the extra placement or sensor embeddings that were concatenated in the other comparisons.

%% file: tables/main_breadth.tex
\begin{table}[H]
\centering
\caption{{\textbf{Complete summary of \ours breadth and coverage.}}}
\vspace{3pt}
\label{tab:inertia1_breadth}
\setlength{\tabcolsep}{6pt}
\renewcommand{\arraystretch}{1.1}
{\small
\begin{adjustbox}{width=0.9\textwidth}
\begin{tabularx}{\textwidth}{lY}
\toprule[1.5pt]
\textbf{Component} & \textbf{Coverage} \\
\midrule
\midrule

\textbf{Sensors} &
\bACC; \bGYR; \bMAG \\
\midrule
\textbf{Device Placements} &
\bWR; \bHN; \bAR; \bBK; \bCH; \bHP; \bKN; \bLG; \bSH; \bAN \\
\midrule
\textbf{Sampling Frequency} &
20 Hz; 5 Hz; 1 Hz; 0.2 Hz \\
\midrule
\textbf{Window Length} &
10 s; 30 s; 60 s; 2 h \\
\midrule
\textbf{Representation Domains} &
Time; Time–Frequency \\
\midrule
\textbf{Model Families} &
GRU; Transformer; ViT (time); ViT (spectrogram); wave2vec CNN; wave2vec Transformer \\
\midrule
\textbf{Pretraining Objectives} &
\begin{tabular}[t]{@{}l@{}}
\textbf{\textit{Autoregressive}}: \argru, \artransformer \\
\textbf{\textit{Contrastive}}: \relcon, \sslwearables, \simclr \\
\textbf{\textit{Self-Distillation}}: \dino \\
\textbf{\textit{Masked Reconstruction}}: \patchtst, \selfpab, \lsm
\end{tabular} \\
\midrule
\textbf{Model Scale} &
Small ($\sim$5M); Medium ($\sim$30M); Large ($\sim$100M) \\
\midrule
\textbf{Downstream Tasks} &
\har; \fog; \dd \\
\midrule
\textbf{Evaluation Protocol} &
Linear Probing; Full Fine-tuning \\
\bottomrule[1.5pt]
\end{tabularx}
\end{adjustbox}
}
\vspace{-5pt}
\end{table}

%% file: tables/appendix_downstream_data.tex
\begin{table}[ht]
\centering
\caption{{
\textbf{Summary of the datasets used in \ours.}} Train and test samples are reported in subjects for \nhanes and \ukb, 30 second windows for \har and \fog downstream datasets, and subject-days for the \dd tasks.}
\vspace{3pt}
\label{tab:appendix:data}
\setlength{\tabcolsep}{4pt}
\renewcommand{\arraystretch}{1.2}
{\small
\adjustbox{max width=\textwidth}{
\begin{tabular}{llcccccll}
\toprule[1.5pt]
\textbf{Dataset} & \textbf{Task} & \textbf{\# Class} & \textbf{\# Train} & \textbf{\# Val} & \textbf{\# Test} & \textbf{Rate} & \textbf{Sensors} & \textbf{Placement} \\
\midrule
\midrule
\rowcolor{gray!15}
\multicolumn{9}{l}{\emph{\textbf{Pretraining:}}}\\
\midrule
\nhanesCite & Pretraining & $-$ & 11.7K & 1.5K & 1.5K & 1-20 Hz & \bACC & \bWR \\
\ukbCite & Pretraining & $-$ & 80K & $-$ & 20K & 0.2 Hz & \bACC & \bWR \\
\midrule
\rowcolor{gray!15}
\multicolumn{9}{l}{\emph{\textbf{Downstream:}}}\\
\midrule
\captureCite  & \har & 10 & 214.8K & 46.9K & 45.8K & 30 Hz & \bACC & \bWR \\
\harplusCite & \har & 5 & 27.3K & 5.6K & 6.0K & 50 Hz & \bACC & \bCH \bLG \\
\harthCite & \har & 9 & 86.9K & 18.2K & 18.0K & 50 Hz & \bACC & \bCH \bLG \\
\hharCite & \har & 6 & 10.4K & 2.2K & 2.2K & 30 Hz & \bACC \bGYR & \bWR \\
\mhealthCite & \har & 12 & 2.8K & 2.0K & 2.0K & 50 Hz & \bACC \bGYR \bMAG & \bAR \bCH \bAN \\
\opportunityCite & \har & 4 & 25.6K & 12.2K & 9.8K & 30 Hz & \bACC \bGYR \bMAG & \bWR \bHN \bAR \bBK \bHP \bKN \\
\pamapCite & \har & 12 & 16.0K & 9.4K & 13.5K & 100 Hz & \bACC \bGYR \bMAG & \bHN \bCH \bAN \\
\recofitCite & \har & 22 & 63.1K & 13.6K & 13.4K & 50 Hz & \bACC \bGYR & \bAR \\
\wisdmCite & \har & 18 & 111.8K & 24.6K & 25.1K & 20 Hz & \bACC \bGYR & \bWR \\
\wearCite & \har & 18 & 29.0K & 6.7K & 6.5K & 50 Hz & \bACC & \bAR \bLG \\
\daphnetCite & \fog & 2 & 12.5K & 2.6K & 2.8K & 64 Hz & \bACC & \bHP \bLG \bAN \\
\odayfogCite & \fog & 2 & 1.7K & .8K & .8K & 128 Hz & \bACC \bGYR & \bWR \bCH \bHP \bLG \bAN \\
\fogturningCite & \fog & 2 & 4.5K & 1.0K & 1.0K & 128 Hz & \bACC \bGYR & \bSH \\
\nhanesCite (Sleep) & Disease & 2 & 1525 & 207 & 207 & 20 Hz & \bACC & \bWR \\
\nhanesCite (Depression) & Disease & 2 & 1764 & 168 & 221 & 20 Hz & \bACC & \bWR \\
\nhanesCite (Dep. Severity) & Disease & 5 & 2548 & 315 & 289 & 20 Hz & \bACC & \bWR \\
\nhanesCite (Parkinson's) & Disease & 2 & 1267 & 162 & 153 & 20 Hz & \bACC & \bWR \\
\nhanesCite (Diabetes) & Disease & 2 & 1317 & 180 & 151 &  20 Hz & \bACC & \bWR \\
\ukbCite (Osteoarthritis) & Disease & 2 & 89.6K & 19.2K & 19.2K & 0.2 Hz & \bACC & \bWR \\
\ukbCite (Osteoporosis) & Disease & 2 & 27.8K & 6K & 6K & 0.2 Hz & \bACC & \bWR \\
\bottomrule[1.5pt]
\end{tabular}
}}
\end{table}

%% file: tables/appendix_model_sizes.tex
\begin{table}[!t]
\centering
\caption{\small{\textbf{Model sizes in millions of trainable parameters.} Medium and large configurations are reported for best performing SSL methods.}}
\label{tab:appendix_model_sizes}
\vspace{3pt}
\setlength{\tabcolsep}{10pt}
\renewcommand{\arraystretch}{1.15}
{\small
\begin{adjustbox}{width=0.99\textwidth}
\begin{tabularx}{\textwidth}{p{4.0cm} Y Y Y}
\toprule[1.5pt]
\textbf{Model} & \textbf{Small (M)} & \textbf{Medium (M)} & \textbf{Large (M)} \\
\midrule
\artransformer & 8.4 & 33.1 & 100.8 \\
\patchtst       & 8.5 & 33.2 & 100.9 \\
\midrule
\argru         & 3.9 & --   & --    \\
\dino           & 8.1 & --   & --   \\
\simclr         & 5.8 & --   & --    \\
\relcon         & 3.6 & -- & -- \\
\sslwearables & 10.5 & -- \\
\selfpab        & 3.3 & --   & --    \\
Supervised CNN            & 5.0 & --   & --    \\
Supervised ViT           & 5.4 & --   & --    \\
\bottomrule[1.5pt]
\end{tabularx}
\end{adjustbox}
}
\vspace{-7pt}
\end{table}

%% file: tables/appendix-modelcomp-auprc.tex
\begin{table}[t]
\centering
\begin{minipage}[t]{0.752\textwidth}
\vspace{0pt}
\centering
\parbox[t][3em][t]{\linewidth}{
\captionof{table}{\small{\textbf{Full results for AUPRC on \har tasks with both linear probing and full finetuning results.}}}
\label{tab:har_auprc}
}
\small
\setlength{\tabcolsep}{4pt}
\renewcommand{\arraystretch}{1.05}
\begin{adjustbox}{width=1\textwidth,center}
\begin{tabular}{lcccccccccc}
\toprule[1.5pt]
Model & \hhar & \mhealth & \opportunity & \pamap & \recofit & \capture & \harplus & \harth & \wear & \wisdm \\
\midrule
\midrule
\rowcolor{gray!15}
\multicolumn{11}{l}{\emph{Supervised Models}}\\
\addlinespace[1pt]
ViT & 60.2 & 70.7 & 38.1 & 59.5 & 80.1 & 55.3 & 58.7 & 82.2 & 86.2 & 66.5 \\
CNN & 76.1 & 69.9 & 34.9 & 76.2 & 76.2 & 50.7 & 57.2 & 87.2 & 88.1 & 74.7 \\
\midrule
\multicolumn{11}{c}{\textbf{\textit{Linear Probing}}}\\
\addlinespace[1pt]
\midrule
\rowcolor{gray!15}
\multicolumn{11}{l}{\emph{General Pretraining Methods}}\\
\addlinespace[1pt]
\patchtst & \underline{77.3} & \textbf{86.8} & 70.8 & \textbf{82.4} & 85.0 & \underline{53.9} & 61.3 & \underline{85.9} & 94.0 & \textbf{77.9} \\
\artransformer & 67.9 & 74.2 & \textbf{79.5} & 73.1 & \textbf{87.8} & \textbf{55.2} & 59.3 & \textbf{86.6} & \textbf{94.8} & 72.8 \\
\argru & 68.0 & \underline{82.1} & 71.6 & 71.1 & \underline{85.7} & 53.2 & 55.9 & 82.4 & \underline{94.0} & 72.1 \\
\dino & 43.5 & 54.2 & 70.1 & 39.4 & 80.6 & 49.9 & 63.1 & 84.6 & 92.5 & 37.4 \\
\simclr & 38.2 & 60.2 & \underline{76.9} & 48.1 & 77.2 & 50.7 & 61.6 & 80.9 & 89.1 & 38.8 \\
\midrule
\rowcolor{gray!15}
\multicolumn{11}{l}{\emph{Domain-specific Methods}}\\
\addlinespace[1pt]
\sslwearables & 64.3 & 57.5 & 42.6 & 65.2 & 44.3 & 39.9 & 61.4 & 56.7 & 42.0 & 51.1 \\
\relcon & 67.1 & 77.0 & 68.3 & 70.0 & 77.5 & 48.0 & \textbf{65.2} & 77.2 & 87.6 & 57.0 \\
\selfpab & \textbf{87.8} & 55.4 & 40.3 & \underline{74.0} & 75.7 & 48.7 & \underline{64.1} & 84.7 & 87.1 & \underline{77.0} \\
\midrule
\multicolumn{11}{c}{\textbf{\textit{Full Fine-tuning}}}\\
\addlinespace[1pt]
\midrule
\rowcolor{gray!15}
\multicolumn{11}{l}{\emph{General Pretraining Methods}}\\
\addlinespace[1pt]
\patchtst & \textbf{86.2} & \underline{89.0} & 62.9 & \textbf{86.8} & 83.6 & \textbf{58.1} & \underline{67.4} & \underline{85.0} & 92.4 & \textbf{85.7} \\
\artransformer & 73.4 & 85.1 & \underline{72.7} & 77.7 & \underline{84.5} & 56.9 & 57.1 & \textbf{88.5} & \textbf{94.9} & \underline{82.0} \\
\argru & 75.0 & 83.6 & 63.9 & \underline{83.0} & 83.4 & 54.4 & 57.9 & 84.4 & \underline{93.1} & 79.7 \\
\dino & 43.5 & 67.6 & 64.5 & 44.8 & 83.8 & \underline{57.0} & 65.4 & 78.7 & 91.0 & 57.1 \\
\simclr & 40.0 & 67.4 & 68.8 & 47.4 & 80.4 & 56.2 & 62.3 & 79.0 & 92.2 & 64.1 \\
\midrule
\rowcolor{gray!15}
\multicolumn{11}{l}{\emph{Domain-specific Methods}}\\
\addlinespace[1pt]
\sslwearables & 78.9 & 81.1 & \textbf{79.1} & 62.4 & 78.9 & 56.4 & 66.5 & 82.6 & 89.2 & 51.8 \\
\relcon & 77.2 & \textbf{90.2} & 60.9 & 82.8 & \textbf{84.5} & 51.9 & \textbf{67.9} & 78.4 & 92.5 & 79.9 \\
\selfpab & \underline{79.3} & 68.7 & 40.7 & 80.6 & 72.0 & 54.2 & 59.8 & 82.1 & 88.5 & 81.2 \\
\bottomrule[1.5pt]
\end{tabular}
\end{adjustbox}
\end{minipage}
\hfill
\begin{minipage}[t]{0.241\textwidth}
\vspace{0pt}
\centering
\parbox[t][3em][t]{\linewidth}{
\captionof{table}{\small{\textbf{Full results for AUPRC on \fog tasks.}}}
\label{tab:fog_auprc}
}
\small
\setlength{\tabcolsep}{2.14pt}
\renewcommand{\arraystretch}{1.05}
\begin{adjustbox}{width=\linewidth}
\begin{tabular}{ccc}
\toprule[1.5pt]
\fogturning & \odayfog & \daphnet \\
\midrule
\midrule
\rowcolor{gray!15}
\multicolumn{3}{l}{\emph{Supervised Models}}\\
\addlinespace[1pt]
39.4 & 72.4 & 59.2 \\
22.6 & 72.4 & 76.8 \\
\midrule
\multicolumn{3}{c}{\textbf{\textit{Linear Probing}}}\\
\addlinespace[1pt]
\midrule
\rowcolor{gray!15}
\multicolumn{3}{l}{\emph{General Pretraining Methods}}\\
\addlinespace[1pt]
61.4 & 66.0 & \underline{87.5} \\
61.3 & 56.5 & 86.2 \\
71.4 & 69.5 & 82.9 \\
37.7 & \textbf{77.8} & 77.6 \\
\textbf{87.8} & 66.4 & \textbf{87.7} \\
\midrule
\rowcolor{gray!15}
\multicolumn{3}{l}{\emph{Domain-specific Methods}}\\
\addlinespace[1pt]
\underline{84.6} & 52.3 & 39.1 \\
76.2 & \underline{77.1} & 79.3 \\
63.3 & 61.0 & 77.3 \\
\midrule
\multicolumn{3}{c}{\textbf{\textit{Full Fine-tuning}}}\\
\addlinespace[1pt]
\midrule
\rowcolor{gray!15}
\multicolumn{3}{l}{\emph{General Pretraining Methods}}\\
\addlinespace[1pt]
78.8 & 70.8 & \underline{89.0} \\
54.5 & 78.9 & 85.0 \\
40.7 & 65.5 & 83.4 \\
69.9 & \textbf{85.3} & 72.5 \\
67.9 & 49.8 & 86.1 \\
\midrule
\rowcolor{gray!15}
\multicolumn{3}{l}{\emph{Domain-specific Methods}}\\
\addlinespace[1pt]
\textbf{96.1} & 57.4 & 85.2 \\
\underline{83.4} & 75.6 & \textbf{90.9} \\
68.4 & \underline{80.4} & 58.3 \\
\bottomrule[1.5pt]
\end{tabular}
\end{adjustbox}
\end{minipage}
\end{table}

%% file: tables/appendix-model-dp.tex
\begin{table}[t]
\centering
\small
\setlength{\tabcolsep}{3pt}
\caption{\small{\textbf{Disease prediction tasks.} We report \textbf{AUPRC} per task. Best results are \textbf{bolded} and second best are \underline{underlined}.}}
\label{tab:disease_auprc}
\vspace{2pt}
\begin{tabular}{lccccc}
\toprule[1.5pt]
Model & Depr. Severity & Depression & Diabetes & Parkinson's & Sleep \\
\midrule
\midrule
\rowcolor{gray!15}
\multicolumn{6}{l}{\emph{Supervised Models}}\\
\addlinespace[1pt]
ViT & 28.3 & 62.6 & 35.2 & 27.0 & 44.5 \\
CNN & 29.8 & 55.4 & 40.5 & 38.4 & 48.6 \\
\midrule
\rowcolor{gray!15}
\multicolumn{6}{l}{\emph{General Pretraining Methods}}\\
\addlinespace[1pt]
\patchtst & 26.6 & 70.7 & 42.9 & 49.8 & 59.1 \\
\artransformer & \underline{35.0} & \textbf{80.1} & 54.6 & 52.3 & 56.5 \\
\argru & 33.7 & 69.9 & 45.2 & 50.1 & 61.7 \\
\dino & 34.5 & \underline{76.2} & \underline{67.9} & 45.1 & \textbf{78.1} \\
\simclr & 32.4 & 71.9 & \textbf{70.9} & 44.9 & \underline{77.3} \\
\midrule
\rowcolor{gray!15}
\multicolumn{6}{l}{\emph{Domain-specific Methods}}\\
\addlinespace[1pt]
\sslwearables & \textbf{37.9} & 60.4 & 59.9 & \textbf{60.4} & 62.7 \\
\relcon & 32.6 & 73.7 & 64.0 & \underline{53.9} & 54.7 \\
\selfpab & 30.2 & 68.8 & 47.1 & 34.5 & 58.9 \\
\bottomrule[1.5pt]
\end{tabular}
\end{table}

%% file: tables/appendix-model-har-ff.tex
\begin{table}[t]
\centering
\begin{minipage}[t]{0.754\textwidth}
\vspace{0pt}
\centering
\parbox[t][3em][t]{\linewidth}{
\captionof{table}{\small{\textbf{Full reesults for AUROC on HAR tasks with both linear probing and full finetuning results.}}}
\label{tab:appendix:har_ff}
}
\small
\setlength{\tabcolsep}{4pt}
\renewcommand{\arraystretch}{1.05}
\begin{adjustbox}{width=1\textwidth,center}
\begin{tabular}{lcccccccccc}
\toprule[1.5pt]
Model & \hhar & \mhealth & \opportunity & \pamap & \recofit & \capture & \harplus & \harth & \wear & \wisdm \\
\midrule
\midrule
\rowcolor{gray!15}
\multicolumn{11}{l}{\emph{Supervised Models}}\\
\addlinespace[1pt]
ViT & 83.9 & 93.8 & 72.9 & 92.0 & 97.7 & 92.9 & 96.6 & 98.7 & 97.7 & 94.1 \\
CNN & 89.1 & 90.7 & 69.5 & 95.5 & 97.0 & 90.9 & 94.5 & 98.7 & 98.0 & 94.8 \\
\midrule
\multicolumn{11}{c}{\textbf{\textit{Linear Probing}}}\\
\addlinespace[1pt]
\midrule
\rowcolor{gray!15}
\multicolumn{11}{l}{\emph{General Pretraining Methods}}\\
\addlinespace[1pt]
\patchtstCite & \underline{93.5} & \textbf{98.2} & 89.8 & \textbf{97.4} & 98.5 & 91.6 & 95.8 & \underline{98.9} & 99.2 & \underline{95.5} \\
\artransformerCite & 88.6 & 96.0 & \textbf{92.1} & \underline{95.6} & \textbf{98.9} & \textbf{92.3} & 95.5 & \textbf{99.0} & \textbf{99.3} & 94.8 \\
\argruCite & 89.2 & 96.0 & 89.4 & 95.0 & \underline{98.7} & 91.3 & 95.3 & 98.5 & \underline{99.2} & 94.7 \\
\dinoCite & 73.4 & 87.2 & 87.8 & 82.7 & 98.1 & 92.0 & 95.4 & 98.9 & 98.5 & 85.2 \\
\simclrCite & 68.6 & 90.0 & \underline{90.0} & 86.8 & 97.8 & \underline{92.2} & \underline{95.9} & 98.6 & 98.0 & 86.4 \\
\midrule
\rowcolor{gray!15}
\multicolumn{11}{l}{\emph{Domain-specific Methods}}\\
\addlinespace[1pt]
\sslwearablesCite & 90.0 & 89.9 & 73.1 & 91.8 & 91.0 & 87.0 & 85.4 & 92.0 & 88.5 & 90.2 \\
\relconCite & 87.8 & \underline{96.3} & 85.3 & 94.5 & 97.0 & 89.9 & \textbf{95.9} & 98.0 & 97.9 & 90.5 \\
\selfpabCite & \textbf{96.8} & 91.8 & 74.7 & 95.4 & 97.4 & 89.7 & 93.4 & 98.6 & 97.7 & \textbf{96.7} \\
\midrule
\multicolumn{11}{c}{\textbf{\textit{Full Fine-tuning}}}\\
\addlinespace[1pt]
\midrule
\rowcolor{gray!15}
\multicolumn{11}{l}{\emph{General Pretraining Methods}}\\
\addlinespace[1pt]
\patchtstCite & \textbf{96.1} & \textbf{98.6} & 87.6 & 97.1 & \underline{98.1} & 93.0 & \textbf{96.9} & 98.3 & 98.8 & \underline{97.4} \\
\artransformerCite & 90.9 & 97.8 & \underline{90.2} & \underline{97.2} & 98.0 & 93.1 & 94.8 & \textbf{99.0} & \textbf{99.1} & 97.3 \\
\argruCite & 92.4 & 97.7 & 88.2 & \textbf{97.4} & \textbf{98.2} & 92.5 & \underline{95.6} & \underline{98.8} & \underline{98.9} & 96.6 \\
\dinoCite & 71.1 & 93.1 & 84.7 & 84.4 & 96.9 & 93.1 & 90.9 & 94.0 & 97.6 & 92.1 \\
\simclrCite & 62.4 & 89.4 & 84.2 & 79.0 & 95.8 & \underline{93.2} & 91.2 & 94.6 & 97.0 & 91.3 \\
\midrule
\rowcolor{gray!15}
\multicolumn{11}{l}{\emph{Domain-specific Methods}}\\
\addlinespace[1pt]
\sslwearablesCite & \underline{94.8} & 96.1 & \textbf{91.6} & 90.3 & 98.1 & \textbf{93.4} & 92.7 & 98.6 & 97.9 & 90.8 \\
\relconCite & 85.8 & \underline{98.4} & 86.4 & 95.3 & 96.5 & 89.8 & 76.5 & 95.5 & 97.4 & 94.0 \\
\selfpabCite & 92.9 & 94.4 & 67.3 & 97.0 & 97.4 & 91.5 & 93.4 & 98.5 & 97.8 & \textbf{97.4} \\
\bottomrule[1.5pt]
\end{tabular}
\end{adjustbox}
\end{minipage}
\hfill
\begin{minipage}[t]{0.239\textwidth}
\vspace{0pt}
\centering
\parbox[t][3em][t]{\linewidth}{
\captionof{table}{\small{\textbf{Full results for AUROC on FOG tasks.}}}
\label{tab:appendix:fog_ff}
}
\small
\setlength{\tabcolsep}{3pt}
\renewcommand{\arraystretch}{1.05}
\begin{adjustbox}{width=\linewidth}
\begin{tabular}{ccc}
\toprule[1.5pt]
\fogturning & \odayfog & \daphnet \\
\midrule
\midrule

\rowcolor{gray!15}
\multicolumn{3}{l}{\emph{Supervised Models}}\\
\addlinespace[1pt]

63.6 & 68.0 & 83.5 \\
51.2 & 80.5 & 87.6 \\

\midrule
\multicolumn{3}{c}{\textbf{\textit{Linear Probing}}}\\
\addlinespace[1pt]
\midrule

\rowcolor{gray!15}
\multicolumn{3}{l}{\emph{General Pretraining Methods}}\\
\addlinespace[1pt]

85.7 & 65.4 & \textbf{95.0} \\
80.2 & 58.6 & 94.3 \\
83.0 & 71.6 & 93.4 \\
61.4 & \textbf{83.2} & 89.2 \\
\underline{91.4} & 66.7 & \underline{94.8} \\

\midrule

\rowcolor{gray!15}
\multicolumn{3}{l}{\emph{Domain-specific Methods}}\\
\addlinespace[1pt]

\textbf{91.6} & 62.0 & 71.4 \\
87.5 & \underline{78.2} & 93.0 \\
80.4 & 55.0 & 92.5 \\

\midrule
\multicolumn{3}{c}{\textbf{\textit{Full Fine-tuning}}}\\
\addlinespace[1pt]
\midrule

\rowcolor{gray!15}
\multicolumn{3}{l}{\emph{General Pretraining Methods}}\\
\addlinespace[1pt]

83.9 & 75.0 & \underline{95.6} \\
78.3 & 68.0 & 93.2 \\
77.7 & 65.5 & 93.2 \\
85.5 & \textbf{86.0} & 86.1 \\
89.3 & 51.5 & 93.8 \\

\midrule

\rowcolor{gray!15}
\multicolumn{3}{l}{\emph{Domain-specific Methods}}\\
\addlinespace[1pt]

\textbf{98.2} & 73.3 & 93.6 \\
\underline{89.5} & 81.2 & \textbf{96.6} \\
83.9 & \underline{85.5} & 80.7 \\

\bottomrule[1.5pt]
\end{tabular}
\end{adjustbox}
\end{minipage}
\end{table}

%% file: tables/appendix_ablation_auroc.tex
\begin{table}[t]
\centering
\small
\setlength{\tabcolsep}{3pt}
\caption{\small{\textbf{Ablation study across task families.} We report \textbf{AUROC} per dataset/task across axis settings.}}
\label{tab:ablation_auroc_all}
\vspace{2pt}

\begin{minipage}{\textwidth}
\centering
\textbf{Human Activity Recognition}
\vspace{2pt}

\begin{adjustbox}{width=\textwidth,center}
\begin{tabular}{lllcccccccccc}
\toprule[1.5pt]
Axis & Model & Axis Value & \hhar & \mhealth & \opportunity & \pamap & \recofit & \capture & \harplus & \harth & \wear & \wisdm \\
\midrule
\midrule
\rowcolor{gray!15}
\multicolumn{13}{l}{\emph{Window Size}}\\
\addlinespace[1pt]
 & \artransformer & 10s & 89.7 & 97.1 & 87.2 & 94.7 & 98.6 & 70.7 & 96.0 & 98.9 & 99.4 & 96.6 \\
 & \artransformer & 30s & 90.9 & 97.8 & 90.2 & 97.2 & 98.0 & 93.1 & 94.8 & 99.0 & 99.1 & 97.3 \\
 & \artransformer & 60s & 85.5 & 96.8 & 87.0 & 97.0 & 97.8 & 53.8 & 97.2 & 98.9 & 99.2 & 92.3 \\
& \patchtst & 10s & 93.8 & 96.4 & 88.0 & 95.0 & 98.3 & 72.5 & 95.8 & 98.4 & 99.0 & 96.6 \\
 & \patchtst & 30s & 96.1 & 98.6 & 87.6 & 97.1 & 98.1 & 93.0 & 96.9 & 98.3 & 98.8 & 97.4 \\
 & \patchtst & 60s & 86.2 & 97.5 & 85.8 & 96.0 & 97.6 & 45.5 & 97.8 & 98.1 & 97.8 & 90.4 \\
\midrule
\rowcolor{gray!15}
\multicolumn{13}{l}{\emph{Sampling Frequency}}\\
\addlinespace[1pt]
 & \artransformer & 1Hz & 76.3 & 96.1 & 88.6 & 94.5 & 97.6 & 51.4 & 99.2 & 97.5 & 98.5 & 95.4 \\
 & \artransformer & 5Hz & 84.8 & 97.2 & 89.4 & 95.9 & 98.0 & 72.1 & 96.6 & 97.9 & 98.6 & 97.0 \\
 & \artransformer & 20Hz & 90.9 & 97.8 & 90.2 & 97.2 & 98.0 & 93.1 & 94.8 & 99.0 & 99.1 & 97.3 \\
& \patchtst & 1Hz & 51.3 & 83.0 & 67.3 & 74.9 & 81.1 & 50.2 & 88.4 & 89.5 & 70.6 & 76.0 \\
 & \patchtst & 5Hz & 90.6 & 94.3 & 92.6 & 94.0 & 97.7 & 53.6 & 96.7 & 97.7 & 98.1 & 96.3 \\
 & \patchtst & 20Hz & 96.1 & 98.6 & 87.6 & 97.1 & 98.1 & 93.0 & 96.9 & 98.3 & 98.8 & 97.4 \\
\midrule
\rowcolor{gray!15}
\multicolumn{13}{l}{\emph{Sensor Axes}}\\
\addlinespace[1pt]
 & \artransformer & Uniaxial & 93.1 & 95.5 & 84.7 & 95.8 & 96.4 & 62.9 & 91.4 & 94.2 & 96.4 & 95.3 \\
 & \artransformer & Triaxial & 90.9 & 97.8 & 90.2 & 97.2 & 98.0 & 93.1 & 94.8 & 99.0 & 99.1 & 97.3 \\
 & \patchtst & Uniaxial & 71.8 & 76.5 & 54.9 & 72.6 & 86.6 & 49.0 & 54.8 & 85.5 & 70.8 & 84.2 \\
 & \patchtst & Triaxial & 96.1 & 98.6 & 87.6 & 97.1 & 98.1 & 93.0 & 96.9 & 98.3 & 98.8 & 97.4 \\
\bottomrule[1.5pt]
\end{tabular}
\end{adjustbox}
\end{minipage}

\vspace{8pt}

\begin{minipage}[t]{0.42\textwidth}
\centering
\textbf{Freezing of Gait}
\vspace{2pt}

\begin{adjustbox}{width=\textwidth,center}
\begin{tabular}{llccc}
\toprule[1.5pt]
Model & Axis Value & \fogturning & \odayfog & \daphnet\\
\midrule
\midrule
\rowcolor{gray!15}
\multicolumn{5}{l}{\emph{Window Size}}\\
\addlinespace[1pt]
\artransformer & 10s & 79.9 & 68.5 & 93.2 \\
\artransformer & 30s & 78.3 & 68.0 & 93.2 \\
\artransformer & 60s & 94.4 & 82.6 & 95.8 \\
\patchtst & 10s & 81.6 & 57.5 & 88.9 \\
\patchtst & 30s & 83.9 & 75.0 & 95.6 \\
\patchtst & 60s & 97.6 & 98.3 & 92.4 \\
\midrule
\rowcolor{gray!15}
\multicolumn{5}{l}{\emph{Sampling Frequency}}\\
\addlinespace[1pt]
\artransformer & 1Hz & 62.2 & 67.4 & 77.8 \\
\artransformer & 5Hz & 91.4 & 68.4 & 86.2 \\
\artransformer & 20Hz & 78.3 & 68.0 & 93.2 \\
\patchtst & 1Hz & 56.0 & 68.5 & 59.6 \\
\patchtst & 5Hz & 71.2 & 65.6 & 84.7 \\
\patchtst & 20Hz & 83.9 & 75.0 & 95.6 \\
\midrule
\rowcolor{gray!15}
\multicolumn{5}{l}{\emph{Sensor Axes}}\\
\addlinespace[1pt]
\artransformer & Uniaxial & 64.0 & 78.0 & 87.9 \\
\artransformer & Triaxial & 78.3 & 68.0 & 93.2 \\
\patchtst & Uniaxial & 60.7 & 53.5 & 50.6 \\
\patchtst & Triaxial & 83.9 & 75.0 & 95.6 \\
\bottomrule[1.5pt]
\end{tabular}
\end{adjustbox}
\end{minipage}
\hfill
\begin{minipage}[t]{0.54\textwidth}
\centering
\textbf{Disease Prediction}
\vspace{2pt}

\begin{adjustbox}{width=\textwidth,center}
\begin{tabular}{llccccc}
\toprule[1.5pt]
Model & Axis Value & Depr. Severity & Depression & Diabetes & Parkinson's & Sleep \\
\midrule
\midrule
\rowcolor{gray!15}
\multicolumn{7}{l}{\emph{Window Size}}\\
\addlinespace[1pt]
\artransformer & 10s & 58.9 & 76.2 & 52.7 & 71.5 & 56.6 \\
\artransformer & 30s & 58.4 & 81.3 & 67.7 & 75.0 & 61.7 \\
\artransformer & 60s & 53.8 & 74.2 & 62.3 & 69.6 & 51.3 \\
\patchtst & 10s & 53.2 & 55.2 & 54.8 & 52.3 & 29.1 \\
\patchtst & 30s & 55.0 & 77.0 & 53.8 & 72.8 & 44.7 \\
\patchtst & 60s & 55.0 & 58.3 & 56.7 & 64.2 & 49.4 \\
\midrule
\rowcolor{gray!15}
\multicolumn{7}{l}{\emph{Sampling Frequency}}\\
\addlinespace[1pt]
\artransformer & 1Hz & 56.6 & 58.2 & 56.1 & 65.0 & 46.7 \\
\artransformer & 5Hz & 56.1 & 72.9 & 54.2 & 68.5 & 32.4 \\
\artransformer & 20Hz & 58.4 & 81.3 & 67.7 & 75.0 & 61.7 \\
\patchtst & 1Hz & 50.7 & 50.7 & 50.4 & 50.4 & 50.0 \\
\patchtst & 5Hz & 54.9 & 62.3 & 49.6 & 60.0 & 27.5 \\
\patchtst & 20Hz & 55.0 & 77.0 & 53.8 & 72.8 & 59.6 \\
\midrule
\rowcolor{gray!15}
\multicolumn{7}{l}{\emph{Sensor Axes}}\\
\addlinespace[1pt]
\artransformer & Uniaxial & 57.8 & 77.6 & 64.6 & 65.8 & 72.7 \\
\artransformer & Triaxial & 58.4 & 81.3 & 67.7 & 75.0 & 61.7 \\
\patchtst & Uniaxial & 51.9 & 55.7 & 63.8 & 51.5 & 54.2 \\
\patchtst & Triaxial & 55.0 & 77.0 & 53.8 & 72.8 & 59.6 \\
\bottomrule[1.5pt]
\end{tabular}
\end{adjustbox}
\end{minipage}

\end{table}

%% file: tables/appendix_ablation_auprc.tex
\begin{table}[t]
\centering
\small
\setlength{\tabcolsep}{3pt}
\caption{\small{\textbf{Ablation study across task families.} We report \textbf{AUPRC} per dataset/task across axis settings.}}
\label{tab:ablation_auprc_all}
\vspace{2pt}

\begin{minipage}{\textwidth}
\centering
\textbf{Human Activity Recognition}
\vspace{2pt}

\begin{adjustbox}{width=\textwidth,center}
\begin{tabular}{lllcccccccccc}
\toprule[1.5pt]
Axis & Model & Axis Value & \hhar & \mhealth & \opportunity & \pamap & \recofit & \capture & \harplus & \harth & \wear & \wisdm \\
\midrule
\midrule
\rowcolor{gray!15}
\multicolumn{13}{l}{\emph{Window Size}}\\
\addlinespace[1pt]
 & \artransformer & 10s & 68.3 & 81.1 & 66.4 & 71.1 & 85.2 & 34.2 & 67.6 & 89.0 & 92.3 & 77.6 \\
 & \artransformer & 30s & 73.4 & 85.1 & 72.7 & 77.7 & 84.5 & 56.9 & 57.1 & 88.5 & 94.9 & 82.0 \\
 & \artransformer & 60s & 57.1 & 77.2 & 69.0 & 76.1 & 81.6 & 22.6 & 77.8 & 83.0 & 92.6 & 68.6 \\
 & \patchtst & 10s & 81.2 & 73.8 & 63.0 & 78.7 & 85.2 & 39.6 & 64.8 & 82.4 & 90.9 & 80.0 \\
 & \patchtst & 30s & 86.2 & 89.0 & 62.9 & 86.8 & 83.6 & 58.1 & 67.4 & 85.0 & 92.4 & 85.7 \\
 & \patchtst & 60s & 59.2 & 84.0 & 65.7 & 78.4 & 83.9 & 25.5 & 78.3 & 80.2 & 86.7 & 69.1 \\
\midrule
\rowcolor{gray!15}
\multicolumn{13}{l}{\emph{Sampling Frequency}}\\
\addlinespace[1pt]
 & \artransformer & 1Hz & 39.7 & 76.1 & 70.6 & 63.7 & 71.3 & 15.8 & 95.5 & 74.6 & 83.7 & 73.3 \\
 & \artransformer & 5Hz & 53.3 & 78.1 & 72.3 & 68.4 & 80.9 & 36.7 & 74.6 & 75.6 & 92.0 & 81.7 \\
 & \artransformer & 20Hz & 73.4 & 85.1 & 72.7 & 77.7 & 84.5 & 56.9 & 57.1 & 88.5 & 94.9 & 82.0 \\
 & \patchtst & 1Hz & 20.5 & 29.6 & 32.7 & 29.8 & 17.7 & 15.3 & 60.7 & 41.7 & 12.6 & 18.3 \\
 & \patchtst & 5Hz & 68.3 & 64.7 & 74.6 & 70.4 & 79.8 & 24.4 & 76.2 & 76.3 & 90.9 & 76.5 \\
 & \patchtst & 20Hz & 86.2 & 89.0 & 62.9 & 86.8 & 83.6 & 58.1 & 67.4 & 85.0 & 92.4 & 85.7 \\
\midrule
\rowcolor{gray!15}
\multicolumn{13}{l}{\emph{Sensor Axes}}\\
\addlinespace[1pt]
 & \artransformer & Uniaxial & 68.6 & 69.1 & 59.9 & 70.4 & 65.1 & 29.3 & 58.0 & 68.5 & 72.2 & 66.5 \\
 & \artransformer & Triaxial & 73.4 & 85.1 & 72.7 & 77.7 & 84.5 & 56.9 & 57.1 & 88.5 & 94.9 & 82.0 \\
 & \patchtst & Uniaxial & 33.1 & 28.5 & 27.4 & 23.8 & 29.2 & 21.5 & 27.9 & 35.2 & 12.5 & 28.1 \\
 & \patchtst & Triaxial & 86.2 & 89.0 & 62.9 & 86.8 & 83.6 & 58.1 & 67.4 & 85.0 & 92.4 & 85.7 \\
\bottomrule[1.5pt]
\end{tabular}
\end{adjustbox}
\end{minipage}

\vspace{8pt}

\begin{minipage}[t]{0.4\textwidth}
\centering
\textbf{Freezing of Gait}
\vspace{2pt}

\begin{adjustbox}{width=\textwidth,center}
\begin{tabular}{llccc}
\toprule[1.5pt]
Model & Axis Value & \fogturning & \odayfog & \daphnet\\
\midrule
\midrule
\rowcolor{gray!15}
\multicolumn{5}{l}{\emph{Window Size}}\\
\addlinespace[1pt]
\artransformer & 10s & 70.2 & 70.8 & 65.3 \\
\artransformer & 30s & 54.5 & 78.9 & 85.0 \\
\artransformer & 60s & 90.3 & 45.7 & 91.3 \\
\patchtst & 10s & 68.8 & 65.0 & 59.2 \\
\patchtst & 30s & 78.8 & 70.8 & 89.0 \\
\patchtst & 60s & 95.0 & 22.1 & 84.6 \\
\midrule
\rowcolor{gray!15}
\multicolumn{5}{l}{\emph{Sampling Frequency}}\\
\addlinespace[1pt]
\artransformer & 1Hz & 33.0 & 64.9 & 53.1 \\
\artransformer & 5Hz & 79.2 & 30.5 & 65.0 \\
\artransformer & 20Hz & 54.5 & 78.9 & 85.0 \\
\patchtst & 1Hz & 28.5 & 55.7 & 34.5 \\
\patchtst & 5Hz & 57.6 & 54.7 & 64.3 \\
\patchtst & 20Hz & 78.8 & 70.8 & 89.0 \\
\midrule
\rowcolor{gray!15}
\multicolumn{5}{l}{\emph{Sensor Axes}}\\
\addlinespace[1pt]
\artransformer & Uniaxial & 41.0 & 78.9 & 75.9 \\
\artransformer & Triaxial & 54.5 & 78.9 & 85.0 \\
\patchtst & Uniaxial & 29.1 & 38.1 & 23.8 \\
patchtst & Triaxial & 78.8 & 70.8 & 89.0 \\
\bottomrule[1.5pt]
\end{tabular}
\end{adjustbox}
\end{minipage}
\hfill
\begin{minipage}[t]{0.55\textwidth}
\centering
\textbf{Disease Prediction}
\vspace{2pt}

\begin{adjustbox}{width=\textwidth,center}
\begin{tabular}{lllccccc}
\toprule[1.5pt]
Axis & Model & Axis Value & Depr. Severity & Depression & Diabetes & Parkinson's & Sleep \\
\midrule
\midrule
\rowcolor{gray!15}
\multicolumn{8}{l}{\emph{Window Size}}\\
\addlinespace[1pt]
 & \artransformer & 10s & 35.0 & 72.5 & 48.0 & 40.0 & 56.3 \\
 & \artransformer & 30s & 35.0 & 80.1 & 54.6 & 52.3 & 56.5 \\
 & \artransformer & 60s & 31.4 & 66.6 & 46.8 & 44.9 & 53.4 \\
 & \patchtst & 10s & 27.0 & 59.4 & 35.8 & 28.2 & 36.6 \\
 & \patchtst & 30s & 26.6 & 70.7 & 42.9 & 49.8 & 46.6 \\
 & \patchtst & 60s & 27.6 & 61.4 & 43.4 & 37.7 & 49.9 \\
\midrule
\rowcolor{gray!15}
\multicolumn{8}{l}{\emph{Sampling Frequency}}\\
\addlinespace[1pt]
 & \artransformer & 1Hz & 32.7 & 58.2 & 41.6 & 43.5 & 48.4 \\
 & \artransformer & 5Hz & 31.2 & 67.2 & 37.3 & 43.7 & 41.3 \\
 & \artransformer & 20Hz & 35.0 & 80.1 & 54.6 & 52.3 & 56.5 \\
 & \patchtst & 1Hz & 21.8 & 48.2 & 33.2 & 24.5 & 39.1 \\
 & \patchtst & 5Hz & 29.8 & 61.2 & 37.0 & 36.8 & 36.7 \\
 & \patchtst & 20Hz & 26.6 & 70.7 & 42.9 & 49.8 & 59.1 \\
\midrule
\rowcolor{gray!15}
\multicolumn{8}{l}{\emph{Sensor Axes}}\\
\addlinespace[1pt]
 & \artransformer & Uniaxial & 33.2 & 78.0 & 51.8 & 55.5 & 70.0 \\
 & \artransformer & Triaxial & 35.0 & 80.1 & 54.6 & 52.3 & 56.5 \\
 & \patchtst & Uniaxial & 23.1 & 49.4 & 45.2 & 24.1 & 42.0 \\
 & \patchtst & Triaxial & 26.6 & 70.7 & 42.9 & 49.8 & 59.1 \\
\bottomrule[1.5pt]
\end{tabular}
\end{adjustbox}
\end{minipage}

\end{table}

%% file: tables/appendix_cluster_stats.tex
\begin{table}[t]
\centering
\caption{\small{\textbf{Overall cluster quality metrics on UMAP embeddings}, averaged across datasets. Incorporating additional placements yields the best-separated representations.}}
\vspace{3pt}
\label{tab:appendix_cluster_stats}
\setlength{\tabcolsep}{8pt}
\renewcommand{\arraystretch}{1.1}
{\scriptsize
\begin{adjustbox}{width=0.95\textwidth}
\begin{tabular}{lrrrr}
\toprule[1.5pt]
\textbf{Condition} & \textbf{Silhouette $\uparrow$} & \textbf{kNN Purity $\uparrow$} & \textbf{Calinski-Harabasz $\uparrow$} & \textbf{Davies-Bouldin $\downarrow$} \\
\midrule
\grayrow
All Placements & \textbf{0.3686} & \textbf{0.8086} & \textbf{478.0} & \textbf{1.2991} \\
All Sensors            & \underline{0.0881} & 0.6635 & 135.8 & \underline{3.5703} \\
Default                & 0.0634 & \underline{0.6671} & \underline{171.0} & 5.1265 \\
Noise Control          & -0.0534 & 0.5160 & 107.6 & 22.2653 \\
\bottomrule[1.5pt]
\end{tabular}
\end{adjustbox}
}
\vspace{-7pt}
\end{table}